\documentclass[%
 reprint,
 superscriptaddress,
 amsmath,amssymb,
 aps,
onecolumn,
floatfix,
hidelinks,
]{revtex4-2}
\usepackage{sidecap}
\usepackage{graphicx}
\usepackage{dcolumn}
\usepackage{bm}
\usepackage{comment}
\usepackage[dvipsnames,table]{xcolor}
\usepackage{float}
\usepackage{hhline}
\usepackage{enumerate}
\usepackage{multirow}
\usepackage{array}
\usepackage{xfrac}
\usepackage{xr}
\usepackage{hyperref}
\usepackage{siunitx}

\makeatletter
\def\maketitle{
\@author@finish
\title@column\titleblock@produce
\suppressfloats[t]}
\makeatother

\renewcommand\vec{\mathbf}

\makeatletter

\newcommand*{\addFileDependency}[1]{
\typeout{(#1)}
\@addtofilelist{#1}
\IfFileExists{#1}{}{\typeout{No file #1.}}
}\makeatother

\makeatletter

\renewenvironment{widetext@grid}{%
  \par\ignorespaces
  \setbox\widetext@top\vbox{%
   \vskip15\p@
   \hb@xt@\hsize{%
    \leaders\hrule\hfil
    \vrule\@height6\p@
   }%
   \vskip6\p@
  }%
  \setbox\widetext@bot\hb@xt@\hsize{%
    \vrule\@depth6\p@
    \leaders\hrule\hfil
  }%
  \onecolumngrid
  \let\set@footnotewidth\set@footnotewidth@ii
}{%
  \par
  \twocolumngrid\global\@ignoretrue
  \@endpetrue
}%

\makeatother

\DeclareMathOperator*{\argmax}{arg\,max}

\newcolumntype{C}[1]{>{\centering\let\newline\\\arraybackslash\hspace{0pt}}m{#1}}

\begin{document}


\title{Tree-based Learning for High-Fidelity Prediction of Chaos}
\author{Adam Giammarese}
\email[Corresponding author\\]{amg2889@rit.edu}
\affiliation{School of Mathematics and Statistics, Rochester Institute of Technology, Rochester, NY 14623}
\author{Kamal Rana}
\affiliation{Chester F. Carlson Center for Imaging Science, Rochester Institute of Technology, Rochester, NY 14623}
\author{Erik M. Bollt}
\affiliation{Clarkson Center for Complex Systems Science, Clarkson University, Potsdam, NY 13699}
\affiliation{Department of Electrical and Computer Engineering, Clarkson University, Potsdam, NY 13699}
\author{Nishant Malik}
\affiliation{School of Mathematics and Statistics, Rochester Institute of Technology, Rochester, NY 14623}

\date{\today}

\begin{abstract}
Forecasting chaotic systems using machine learning has become highly sought after due to its potential applications in predicting climate and weather phenomena, stock market indices, and pathological activity in biomedical signals. However, existing solutions, such as neural network-based reservoir computing (RC) and long-short-term memory (LSTM), contain numerous model hyperparameters that must be tuned, often requiring high computational resources and large training datasets. Here, we propose a computationally simpler regression tree ensemble-based technique to predict the temporal evolution of chaotic systems in data-driven environments. Furthermore, we introduce a heuristic procedure to prescribe hyperparameters through automated statistical analysis of training data, which eliminates the need for the user to perform hyperparameter tuning. We investigate the efficacy of the proposed hyperparameter prescription procedure through numerical experiments. Lastly, we demonstrate the state-of-the-art performance of our proposed approach on several benchmark tasks, including the Southern Oscillation Index, a crucial but noisy climate time series with limited samples, to illustrate its effectiveness in real-world settings.

\end{abstract}

\maketitle



\section{Introduction}
Recent advances in machine learning techniques have enabled the prediction of the temporal evolution of chaotic systems in entirely data-driven settings, without requiring prior knowledge of the governing equations \cite{pathak2018model,gauthier2021next,Wu2024}. These \emph{model-free} approaches represent a significant breakthrough in modeling and analyzing complex systems and have transformative implications for various fields of science and technology \cite{pathak2018model,jaeger2004harnessing,gauthier2021next,Wu2024,Zhai2023}. Deep learning techniques such as Recurrent Neural Networks (RNN) and, more specifically, Long Short-Term Memory (LSTM) provide substantial performance in forecasting chaotic systems, likely due to their ability to capture fading short-term memory \cite{rumelhart1986learning,elsworth2020time,hochreiter1997long,jaeger2004harnessing,maass2002real}. However, RNN and LSTM are computationally expensive models, and require both the tuning of hyperparameters and a significant amount of training data to provide accurate forecasts. Reservoir computing (RC) reduces computational expense and eliminates the need for large training data sets by randomizing the recurrent pool of nodes within an RNN (called the reservoir), which reduces training to a linear optimization problem \cite{jaeger2004harnessing, maass2002real}. Despite RC's improvements to RNN and LSTM, the broader acceptance of these model-free approaches remains challenging, since their real-world implementation requires tuning many hyperparameters. This computationally intensive process also introduces subjective choices in modeling.

The recently developed Next Generation Reservoir Computing (NG-RC) converts RC into a mathematically equivalent nonlinear vector autoregression (NVAR) machine, which further improves RC by reducing the number of necessary hyperparameters and removing the need for a randomized reservoir \cite{gauthier2021next, bollt2021explaining}. The NG-RC methodology, while promising, does not eliminate the need for hyperparameter tuning, rendering it unappealing for automated applications. Furthermore, while the nonlinear readout of NG-RC allows for accurate prediction of chaotic data, it may become prohibitive when the supplied training data has a high spatial dimension. Additionally, NG-RC and the aforementioned methods do not provide an optimal selection of time delays and lag intervals, which are essential for accurate forecasting of chaotic data but expensive to tune if repeated model training is required (such as in a hyperparameter grid search). Recent advances in RC hyperparameter tuning (and machine learning models in general), such as Bayesian optimization, substantially improve the stability and effectiveness of hyperparameter tuning \cite{yperman2016bayesian, mwamsojo2024stochastic, wu2019hyperparameter}. However, such improvements do not completely eliminate the necessity of hyperparameter tuning in practice, motivating the need for a high-fidelity alternative that does not require hyperparameter tuning.

Tree-based models are extensively used in machine learning due to their remarkable adaptability, versatility, and robustness. These models require only a limited number of hyperparameters, making them an ideal choice for many applications \cite{hastie_09_elements,randomforests}. Furthermore, recent work that has generated much excitement in machine learning communities demonstrates the benefits of tree-based machine learning models, such as XGBoost and Random Forests (RF), over deep learning models on tabular data, both in terms of accuracy and computational resources \cite{chen2016xgboost, randomforests, shwartz2022tabular, grinsztajn2022tree}. Inspired by these recent advances, we investigate tree-based regression as an alternative to the above-mentioned neural network approaches to develop a low complexity yet high-fidelity method for forecasting chaotic data without the need for hyperparameter tuning.

In this work, we introduce a tree-based alternative for learning chaos, eliminating the need for hyperparameter tuning. Furthermore, we present a detailed analysis of the efficacy of this novel approach and show that it outperforms existing methods in accuracy, user-friendliness, and computational simplicity. We name this tree-based method TreeDOX: \textbf{Tree}-based \textbf{D}elay \textbf{O}verembedded e\textbf{X}plicit memory learning of chaos. TreeDOX mimics the implicit fading short-term memory of RNN, LSTM, RC, and NG-RC via explicit short-term memory in the form of time delay overembedding. This overembedding differs from the traditional time-delay embedding due to its intentional usage of a higher embedding dimension than that suggested by Takens' theorem, which helps to model nonstationary dynamical systems \cite{hegger2000coping, verdes2006overembedding, takens2006detecting,malik2020}. TreeDOX uses an ensemble tree method called Extra Trees Regression (ETR) and uses the inherent Gini feature importance of ETR to perform feature reduction on the time delay overembedding \cite{extratrees}, reducing computational resource usage and improving generalizability. We demonstrate the efficacy of TreeDOX on several chaotic systems, including the H\~enon map, Lorenz and Kuramoto-Sivashinsky systems, and a real-world chaotic dataset: the Southern Oscillation Index (SOI).

\begin{figure}
    \centering
    \includegraphics[width=\columnwidth,keepaspectratio]{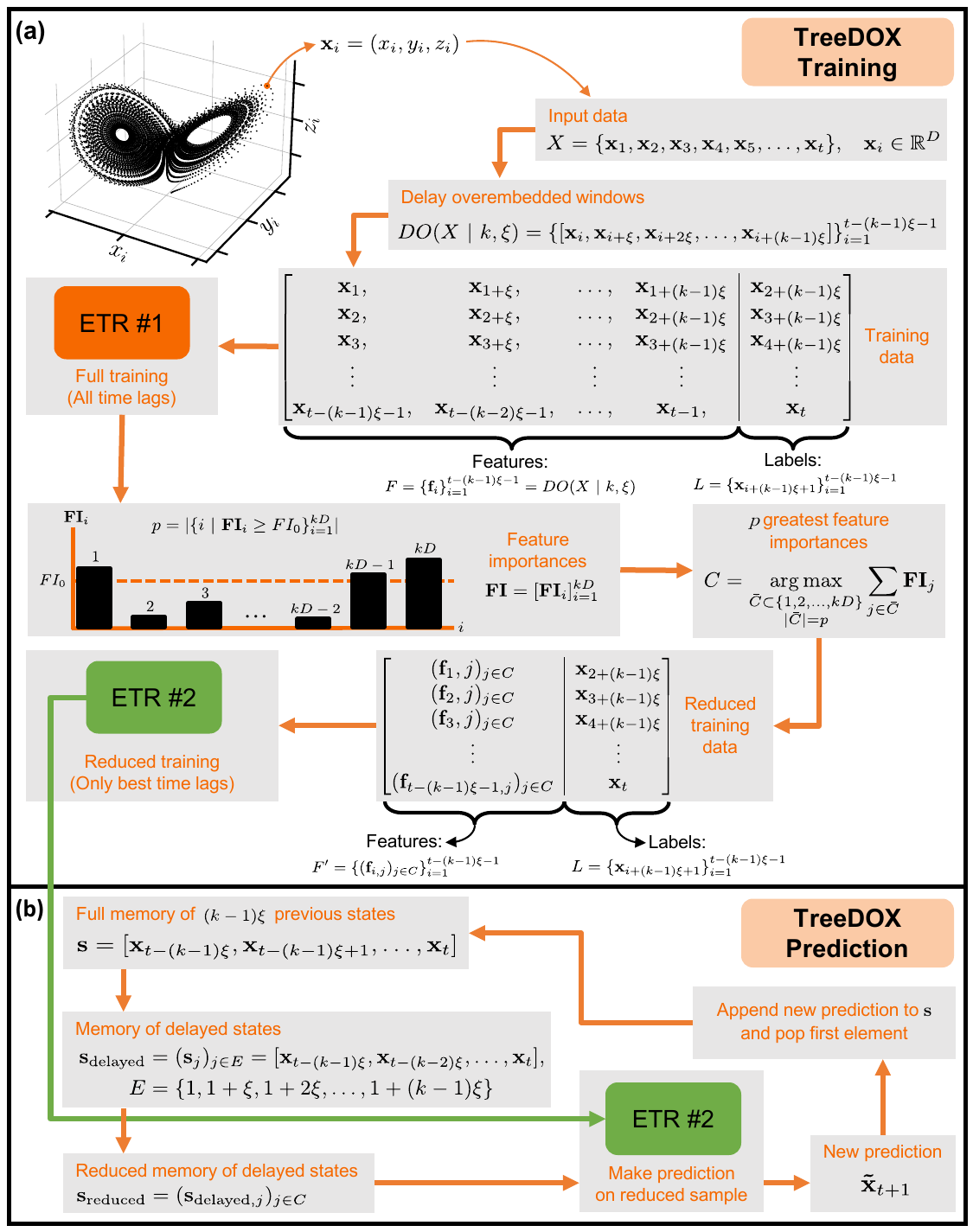}
    \caption{Summary of TreeDOX training \textbf{(a)} and self-evolved prediction \textbf{(b)}. Here, $\vec{x}_i\in\mathbb{R}^D$ represents each input sample, $k$ is the time delay overmbedding dimension, and $\xi$ is the embedding time lag. When running TreeDOX in batched prediction mode (such as when performing `lead' predictions where, at each iteration, the model is given train/test data and is requested to make a prediction a given number of time points ahead), the feedback loop may be ignored. Instead, a batch of respective $\vec{s}_\text{reduced}$ vectors is constructed in advance and fed into the model simultaneously.}
    \label{fig:TreeDOX}
\end{figure}

\section{Results}
Here, we describe the inner workings of TreeDOX (including training, inference, and hyperparameter prescription) and demonstrate its efficacy on a variety of test cases, the first few of which use toy models to validate performance. Firstly, we test TreeDOX on a discrete chaotic system: the H\'enon map. We show that TreeDOX can recreate the H\'enon chaotic attractor, verifying the ability of TreeDOX to capture long-term dynamics in general. Next, we apply TreeDOX to a prototypical continuous chaotic system: the Lorenz system. Aside from showing its ability to recreate the respective chaotic attractor, we show that TreeDOX can make short-term, self-evolved forecasts with a level of accuracy similar to state-of-the-art machine learning methods. The last toy model, the Kuramoto-Sivashinsky system, is used to demonstrate TreeDOX's ability to understand data with high spatial dimension, where the respective data uses $D=64$ dimensions to discretize the spatial domain. Next, we utilize real-world data to demonstrate TreeDOX's effectiveness in a real-world scenario, specifically the open-loop prediction of a given number of time steps in advance. For this purpose, we use the Southern Oscillation Index (SOI) data due to its level of noise and lack of large data volume, unlike the above toy models \cite{ropelewski1987extension, soi-data1}. We demonstrate that for an array of lead values (i.e., the number of steps in advance to make predictions), the accuracy and speed of TreeDOX rival those of its neural network-based competitors, and also that TreeDOX requires less training data. Lastly, we vary the few hyperparameters introduced in the method and make SOI predictions to demonstrate not only the low sensitivity of TreeDOX to these hyperparameters but also that our proposed hyperparameter prescription schemes achieve the best possible accuracy out of TreeDOX while avoiding further diminishing returns in computational speed. Note that in all results that compare TreeDOX with current methods, the methods feature hyperparameter tuning via Tree-structure Parzen Estimators (TPE), a Bayesian optimization scheme \cite{bergstra2011algorithms}. 25 samples are used when performing the TPE hyperparameter tuning.

\subsection{Rationale behind Extra Trees}
While TreeDOX could certainly use the recently popular XGBoost form of tree-based regression, we opt instead for the classical Random Forest due to its typically successful performance under default hyperparameter values, such as the number of trees in the ensemble \cite{oshiro2012many, probst2019hyperparameters, fernandez2014we}. TreeDOX uses a variant of Random Forests called an Extra-Trees Regression (ETR) \cite{extratrees}. The ETR algorithm is an ensemble-based learning method in which each decision tree is constructed using all training samples of the data, rather than bootstrapping. Unlike the standard Random Forest method, where all the attributes are used to determine the locally optimal split of a node, a random subset of features is selected for a node split in an ETR \cite{extratrees}. For each feature in the random subset, a set of random split values is generated within the range of the training data for the given feature, and a loss function (usually mean square error) is calculated for each random split. The feature-split pair that results in the lowest value of the given loss function is selected as the node. For every testing sample, each tree predicts the output value independently, and the final output value of the ETR is the mean of all predictions of the trees. Here, the regression trees that comprise each predictor in the ensemble can understand nonlinear data due to their nature as piecewise constant predictors. Therefore, with enough depth, regression trees can predict nonlinear data with approximate smoothness. Ensembles of sufficiently many regression trees, such as ETR discussed here, can predict nonlinear data more smoothly than individual trees via averaging over the `stepping' effect of the piecewise constant response of the trees.

Apart from calculating testing accuracy, ETRs also quantify the importance of each feature using mean decrease impurity, also known as Gini importance. Another advantage of ETR is its resilience to correlated features and the value of hyperparameters, such as the number of trees in the ensemble, minimum samples per leaf, or max depth of the individual trees. Often, the prescribed hyperparameters will achieve an acceptable regression; instead, the impact of such hyperparameters is felt in the space and time complexity of the algorithm, as training complexity scales linearly with both the number of trees and the number of variables encountered during splitting. ETRs have two advantages over RFs: (1) lower time complexity and variance due to the randomized splits and (2) lower bias due to the lack of bootstrapping. While ETRs are more expensive to train compared to RC, they do not require an expensive grid search to find hyperparameters and tend to be just as fast when making predictions. However, unlike RC and other state-of-the-art methods, ETRs do not contain explicit memory of system variables. Instead, the use of delay overembedding provides explicit memory to the model.

\subsection{TreeDOX Training and Prediction}
TreeDOX uses two ETRs---one whose role is to calculate feature importances and another whose role is to perform predictions using reduced features. Before we formally introduce TreeDOX, we will first define key concepts. First, assume that $D$-dimensional spatiotemporal data is in the following form: ${\displaystyle X = \{\vec{x}_i\}_{i=1}^{t}}$ where $\vec{x}_i \in \mathbb{R}^D$ and $t$ is the length of the temporal component. A time delay overembedding of $X$ will be denoted in the following manner: {${DO(X ~|~ k,\xi) = \{[\vec{x}_i, \vec{x}_{i+\xi}, \dots, \vec{x}_{i+(k-1)\xi}]\}_{i=1}^{t-(k-1)\xi-1}}$} where $k\in\mathbb{N}$ is the overembedding dimension and $\xi\in\mathbb{N}$ is the time lag between observations in the time delay overembedding. We suggest a procedure to prescribe the values of $k$ and $\xi$ below. We start by constructing the set of features and labels used in training, denoted by ${F = \{\vec{f}_i\}_{i=1}^{t-(k-1)\xi-1} = DO(X~|~k,\xi)}$ and ${L = \{\vec{l}_i\}_{i=1}^{t-(k-1)\xi-1} = \{\vec{x}_{i+(k-1)\xi+1}\}_{i=1}^{t-(k-1)\xi-1}}$, respectively. After training ETR \#1 on $(F,L)$ we request the feature importances $\vec{FI} = \{\vec{FI}_j\}_{j=1}^{kD}$ representing the mean Gini importance (across all trees in the ensemble) of the $kD$ columns in the flattened time delay overembedding features, which captures an essence of predictive ability for each time lag. We introduce one hyperparameter, $p\in\mathbb{N}$, whose prescription is described below. We construct a set ${C=\argmax_{\bar{C}} \sum_{j\in\bar{C}} FI_j}$, where $\bar{C} \subset \{1,2,\dots,kD\}$ and $|\bar{C}|=p$. In other words, $C$ contains the indices of the greatest $p$ entries in $\vec{FI}$, and represents the columns of $F$ we wish to use in final training since ETR \#1 finds their respective time delays to hold the most predictive power. Next we construct a new set of reduced features, $F' = \{(\vec{f}_{i,j})_{j\in C}\}_{i=1}^{t-(k-1)\xi-1}$, where $\vec{f}_{i,j}$ represents the $j$-th column in the row vector $\vec{f}_i$. Lastly, we train ETR \#2 on $(F',L)$, which benefits from faster training and increased generalizability due to the feature reduction.

At the beginning of the forecasting stage of TreeDOX, a vector ${\vec{s} = [\vec{x}_{t-(k-1)\xi}, \vec{x}_{t-(k-1)\xi+1}, \dots, \vec{x}_{t}]}$ is initialized from the end of the training data and another vector ${\vec{s}_{\text{delayed}} = (\vec{s}_j)_{j\in E} = [\vec{x}_{t-(k-1)\xi}, \vec{x}_{t-(k-2)\xi}, \dots, \vec{x}_{t}]}$ is collected from ${\vec{s}}$, where ${E=\{1, 1+\xi, \dots, 1+(k-1)\xi\}}$. ${\vec{s}_{\text{reduced}} = (\vec{s}_{\text{delayed},j})_{j\in C}}$ is treated as a sample for the feature-reduced ETR (ETR \#2) from which a prediction of $\vec{x}_{t+1}$, $\tilde{\vec{x}}_{t+1}$ is extracted. $\vec{s}$ is updated to remove the first element and append the prediction $\tilde{\vec{x}}_{t+1}$ to the end. The updated $\vec{s}$ vector is used to repeat the process, hence the \emph{self-evolutionary} nature of TreeDOX forecasting. Diagrams in Fig.~\ref{fig:TreeDOX} summarize the training and forecasting stages of TreeDOX. Since the training of ETR \#1 uses all available time lags in the delay overembedding and is merely used to estimate the predictive power of each time lag, we suggest limiting the number of trees in the ensemble to reduce training time when using large training sets. However, due to the usually fast training of ETR \#2 as a result of reduced features, it is beneficial to use a greater number of trees to decrease the bias in predictions, and only marginally increase training time.

\subsection{Hyperparameter Selection Procedure}
There exist three hyperparameters introduced in TreeDOX that are not otherwise prescribed by standard ETR implementations: dimension, $k$, and lag, $\xi$, of the delay overembedding, and the number of final features to use, $p$. To remove the responsibility of hyperparameter tuning from the TreeDOX user, we wish to leverage the existing training data to prescribe the values of $k$, $\xi$, and $p$. Here, we propose a heuristic procedure to select hyperparameter values that showed consistent success in the numerical experiments outlined later.

\emph{Delay overembedding hyperparameters ($k$ and $\xi$):} To prescribe values for $k$ and $\xi$, we must investigate their role in the time delay overembedding. Since the time delay overembedding has dimension $k$ and lag $\xi$, one may consider that there are moving windows of length $(k-1)\xi$ (the earliest point in the time delay overembedding) used to train TreeDOX. We use Average Mutual Information (AMI) between the training data and its delayed copy shifted by $\tau$ states to determine how much information about the current state is stored in the previous states. We propose the following procedure: \textbf{(1)} calculate $AMI_i(\tau)$ for $\tau \leq \tau_{max}$, where $AMI_i(\tau)$ is the AMI of the $i$-th dimension of training data and its copy shifted by $\tau$ states and $\tau_{max}$ (say 10\% of the training length) is simply an arbitrary limit to avoid unnecessary calculation---$\tau_{max}$ will roughly correlate to the maximum number of training features the user desires for each data dimension; \textbf{(2)} select a quantile threshold of $AMI_i(\tau)$, such as 0.5---while it may initially appear unintuitive, our suggestion of 0.5 as a quantile of $AMI_i(\tau)$ will allow TreeDOX to avoid arbitrarily large overembedding dimension when a system has slowly fading memory; \textbf{(3)} optionally measure the local maxima of $AMI_i(\tau)$; \textbf{(4)} take $\tau_{i,crit}$ to be the smallest such $\tau$ such that $AMI_i(\tau)$ drops below the calculated quantile; \textbf{(5)} if local maxima were calculated and $\tau_{i,crit}$ is less than the $\tau$ of the first maxima, take $\tau_{i,crit}$ to instead be the first maxima---this will force the time delay overembedding to take the marginally increased system memory given by the gap between the initial $\tau_{i,crit}$ and the first local maxima into account. According to this procedure, $\tau_{i,crit}$ will estimate the time delay at which the system does not contain sufficient memory to have predictive power of the next state. Therefore, we suggest choosing $(k-1)\xi$ to be the maximum value of the set of critical $\tau$'s, so that one may force the time delay overembedding to remain in the region where the system retains information about its next state. See Supplementary Fig.~1 for a visualization of the procedure to calculate $\tau_{i,crit}$. The choice of $\xi$ made to specify a value for $k$ from the prescribed $(k-1)\xi$ is a much simpler one: the smaller $\xi$, the more computational resources needed to train TreeDOX---one may think of $\xi$ as a subsampling hyperparameter, where the larger $\xi$, the less data the model is fed. Thus, if one has a powerful enough computer, one should choose $\xi=1$ and thus $k = \max_{1 \leq i \leq D}(\tau_{i,crit})+1$. Otherwise, choose a large enough $\xi$ such that training is a reasonable task and 
\begin{equation}
    k = \frac{1}{\xi}\max_{1 \leq i \leq D}(\tau_{i,crit}) + 1. \label{k}
\end{equation}

\emph{Feature reduction hyperparameter ($p$):} To prescribe a value for $p$, we next investigate the features fed to the ETR. After training ETR\#1, we request the impurity-based feature importances (FIs), also known as the Gini feature importance, to be calculated, which ranks the $kD$ features according to their respective reduction of the specified loss function (mean square error, in our case). Gini importance is defined on a tree-level basis, meaning each tree in the regression ensemble captures its own estimation of the importance of each feature. Therefore, a convenient and natural extension is to consider the statistical properties of feature importance samples. We propose the following procedure: \textbf{(1)} For each $i\in\{1,2,\dots,kD\}$, perform a one-sided t-test on the samples of $\vec{FI}_i$, where the null hypothesis is $H_0:~\mu\leq0$, and alternative hypothesis is $H_1:\mu>0$, where $\mu$ is the population mean of $\vec{FI}_i$; \textbf{(2)} define a null rate, $FI_0$, to be the minimum $\vec{FI}_i$ that passes the respective t-test with a p-value of 5\% or lower; \textbf{(3)} define $p$ to be the number of indices $i$ such that the sample mean of $\vec{FI}_i$ is greater than or equal to $FI_0$. Additionally, one may require that the feature reduction step does, indeed, reduce features by a significant margin. Therefore, we suggest that if $p>kD/2$ (meaning more than half the possible features remain after reduction) then one should repeat the procedure with $H_0:~\mu\leq\text{median}(\vec{FI})$ and $H_1:~\mu>\text{median}(\vec{FI})$. In all displayed TreeDOX results, we use these suggested methods of prescribing $k$, $\xi$, and $p$. See Supplementary Figs.~1 and 2 for visualizations of hyperparameter prescription in practice.

\subsection{Discrete System}
\begin{SCfigure}[50]
    \centering
    \includegraphics[width=.5\columnwidth]{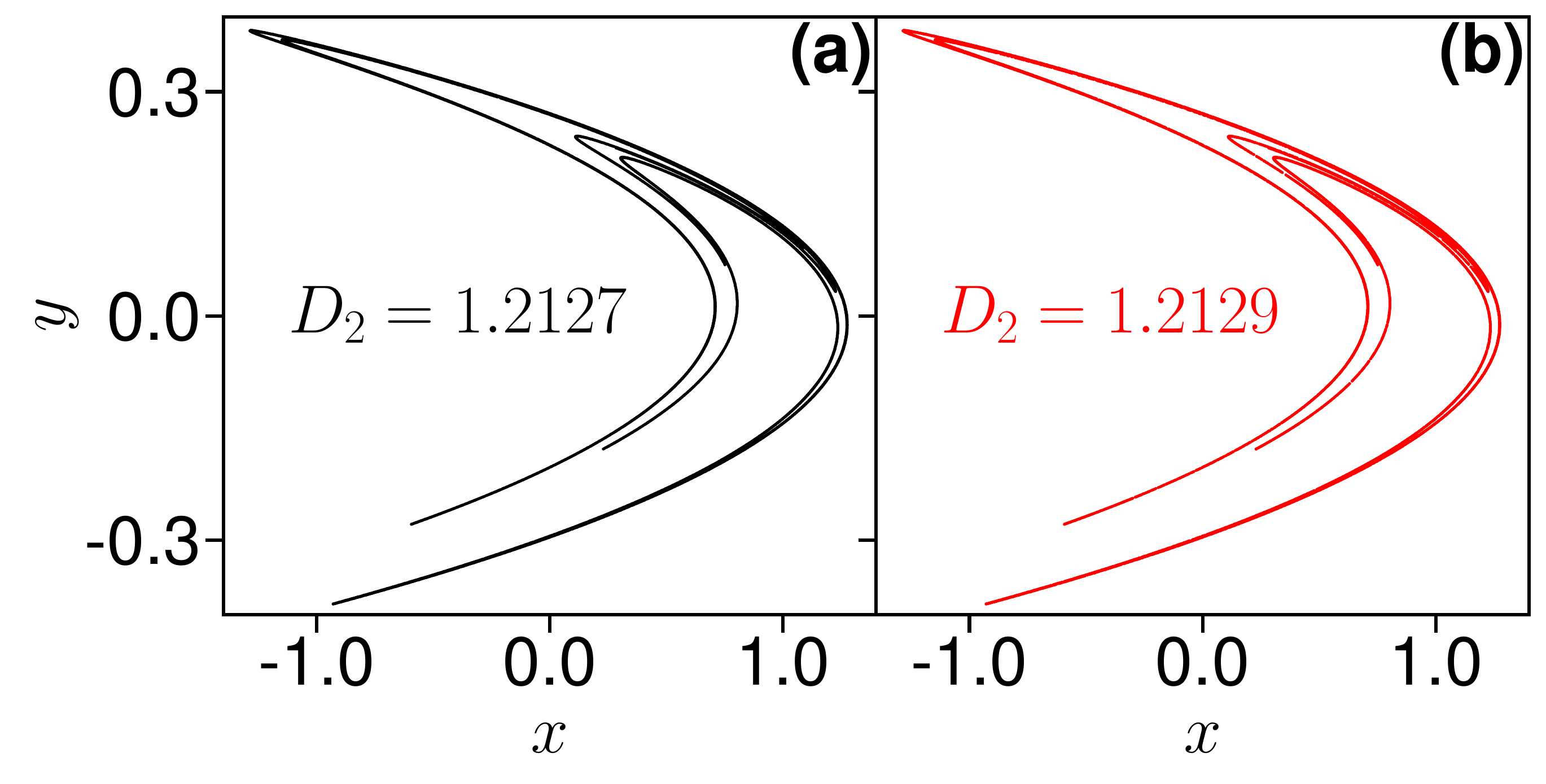}
    \caption{\textbf{(a,b)} Test and predicted H\'{e}non map attractors, respectively, with $a=1.4$ and $b=0.3$. Using $\xi=1$ and a quantile threshold of 0.5, TreeDOX selected $k=16$ according to Eqn.~\ref{k}. Using $\vec{FI}$ from ETR \#1, $p=13$ of $kD=32$ possible features are greater than $FI_0$. $D_2$ is the correlation dimension of the respective attractor. Here, 25,000 and 50,000 training and testing samples were used, respectively.}
    \label{fig:henon}
\end{SCfigure}

As a prototypical discrete chaotic system, the H\'{e}non map is used to verify that TreeDOX can recreate chaotic dynamics. We generate training data from the H\'{e}non map, $(x_{n+1}, y_{n+1}) = (1 - ax_n^2 + y_n, bx_n)$, where $a=1.4$, $b=0.3$, and $(x_0,y_0)=(0,0)$. $25,000$ and $50,000$ training and testing points are used, respectively. Using $\xi=1$ and a quantile threshold of 0.5, TreeDOX selected $k=16$ according to Eqn.~\ref{k}. Using $\vec{FI}$ from ETR \#1, $p=13$ of $kD=32$ possible features are greater than $FI_0$. Fig.~\ref{fig:henon} displays the trajectories of both the test and predicted data. We compare the complexity of the resulting chaotic attractors via correlation dimension, $D_2$ \cite{grassberger1983characterization}, which necessitates a large test set for numerical stability purposes. Supplementary Figs. 3 and 4 show similar results for the logistic map, another archetypal discrete dynamical system. While these findings are not essential for demonstrating the effectiveness of TreeDOX, they do offer valuable insights, including a detailed recreation of the logistic map's bifurcation diagram.

\sidecaptionvpos{figure}{c}

\subsection{Continuous System}
\begin{SCfigure}[50]
    \includegraphics[width=.5\columnwidth,keepaspectratio]{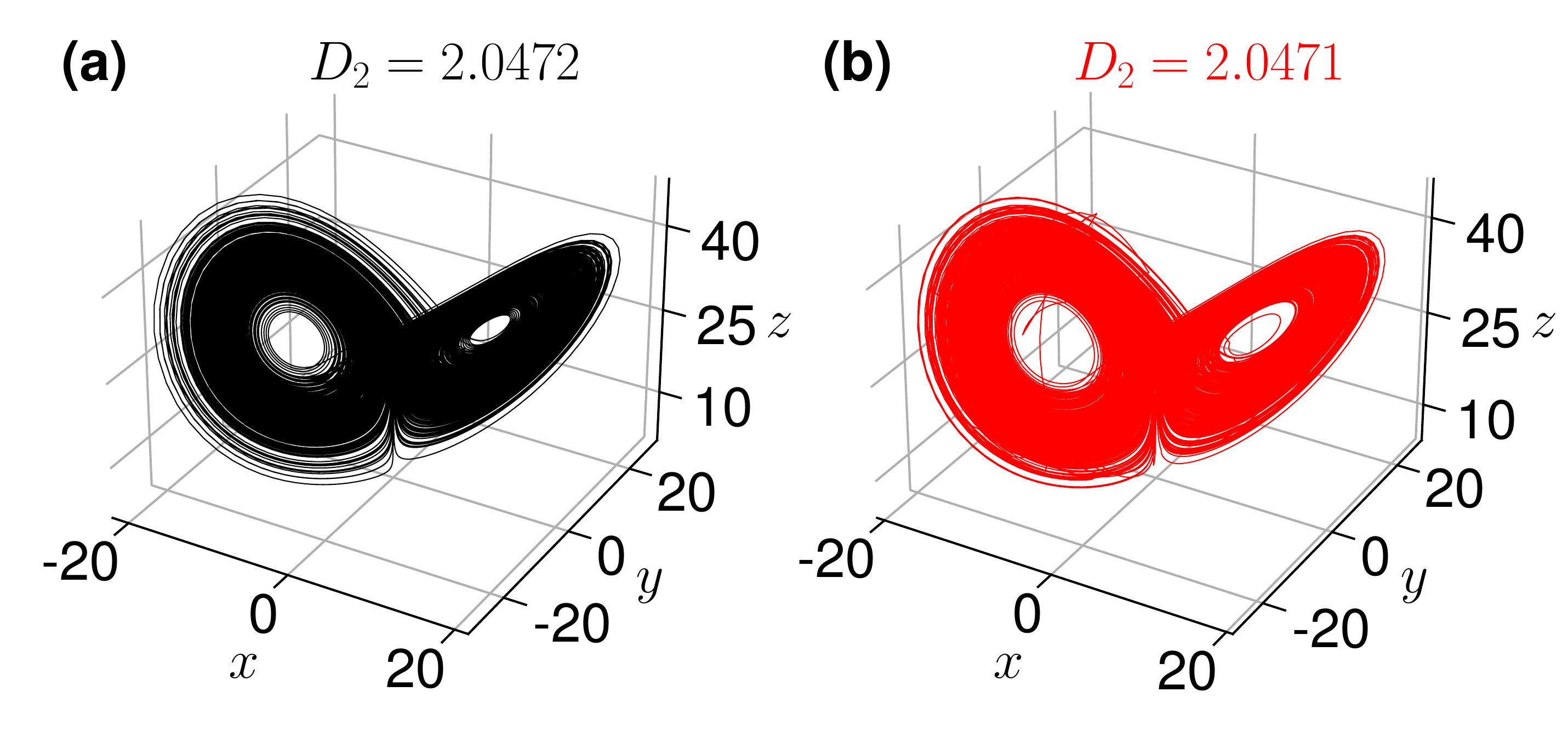}
    \caption{\textbf{(a,b)} Test and predicted Lorenz system attractors, respectively, with $\sigma = 10$, $\beta = 8/3$, and $\rho = 28$. Using $\xi=1$ and a quantile threshold of 0.5, TreeDOX selected $k=109$ according to Eqn.~\ref{k}. Using $\vec{FI}$ from ETR \#1, $p=136$ of $kD=327$ possible features are greater than $FI_0$. $D_2$ is the correlation dimension of the respective attractor. Here, 25,000 and 50,000 training and testing samples were used, respectively.}
    \label{fig:lsattractor}
\end{SCfigure}

\begin{figure}[b]
    \includegraphics[width=\columnwidth,keepaspectratio]{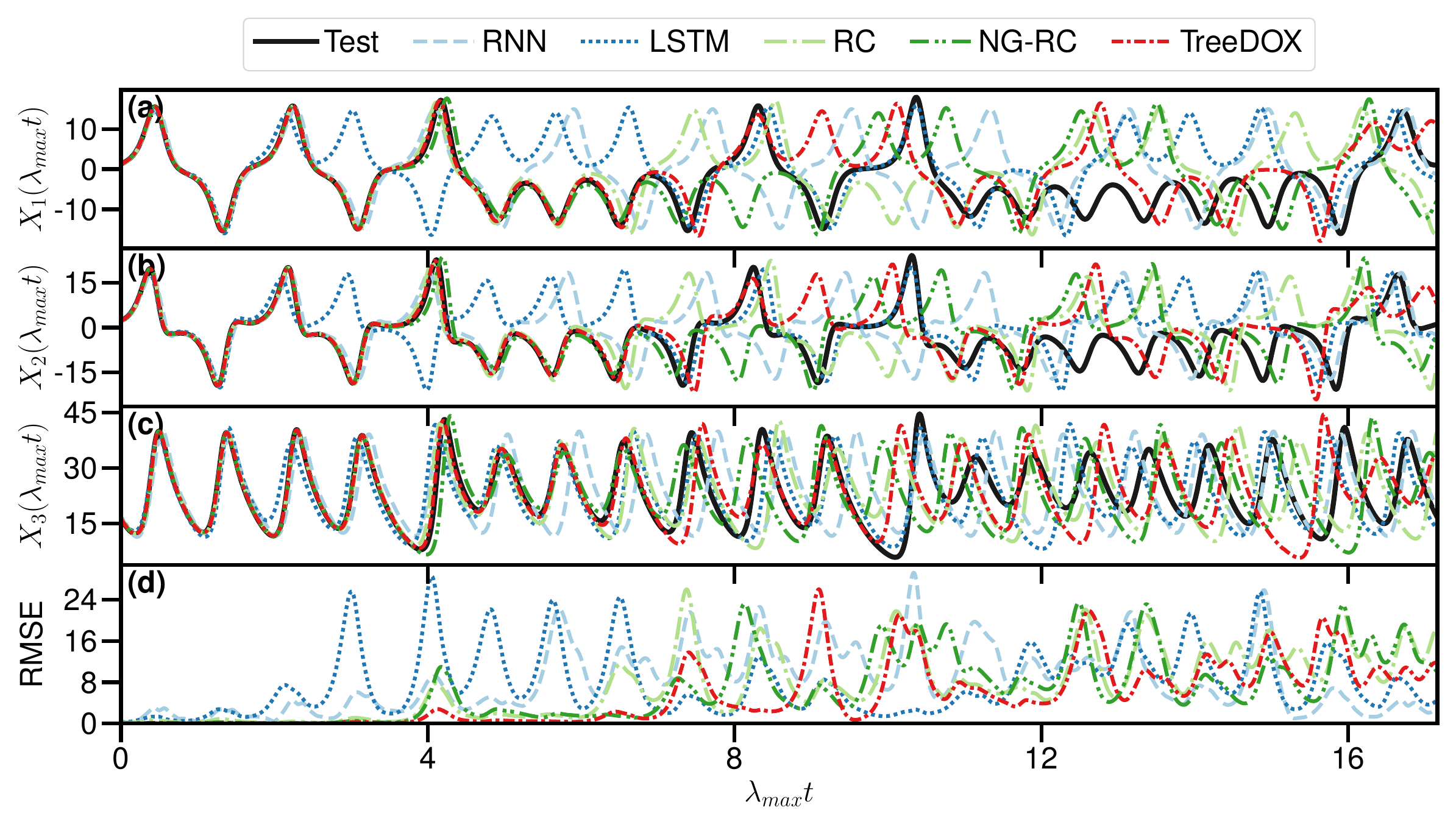}
    \caption{Summary of Lorenz system forecasts with $\sigma = 10$, $\beta = 8/3$, $\rho = 28$, displayed using Lyapunov time, where $\lambda_{max}=0.8739$. Using $\xi=1$ and a quantile threshold of 0.5, TreeDOX selected $k=56$ according to Eqn.~\ref{k}. Using $\vec{FI}$ from ETR \#1, $p=75$ of $kD=168$ possible features are greater than $FI_0$. Test data is black while RNN, LSTM, RC, NG-RC, and TreeDOX are light blue, blue, light green, green, and red, respectively. \textbf{(a,b,c)} $x$, $y$, and $z$ coordinates, respectively. \textbf{(d)} Root Mean Square Error (RMSE) of $x$, $y$ and $z$ forecasted versus test data. Here, 25,000 and 1,500 training and testing samples were used, respectively.}
    \label{fig:lorenz_comparison}
\end{figure}

\begin{sloppypar}
The next benchmark we use is the Lorenz system, a prototypical continuous chaotic system: ${(\dot{x}, \dot{y}, \dot{z}) = (\sigma(y-x), x(\rho-z)-y, xy-\beta z)}$ \cite{lorenz2017deterministic}. Due to multiple nonlinear terms in the generating dynamics, a simple round-off error is enough to cause computational Lorenz system forecasts to diverge quickly, making the numerical forecasting of the system difficult. We select the typical parameters $\sigma = 10$, $\beta = 8/3$, and $\rho = 28$. For both Figs.~\ref{fig:lsattractor} and \ref{fig:lorenz_comparison} RK45 is used to generate training and testing data, where $[x_0, y_0, z_0]=[1,1,1]$ and $dt=0.01$. Fig.~\ref{fig:lsattractor} uses 25,000 and 50,000 training and testing points, respectively, and Fig.~\ref{fig:lorenz_comparison} uses 25,000 and 1,500 training and testing points, respectively. Note that in Fig.~\ref{fig:lorenz_comparison} a greatest Lyapunov exponent of $\lambda_{max}=0.8739$ is assumed \cite{geurts2020lyapunov}. Fig.~\ref{fig:lsattractor} demonstrates TreeDOX's ability to predict long-term dynamics, while Fig.~\ref{fig:lorenz_comparison} portrays TreeDOX's self-evolved forecast accuracy in comparison to current methods.
\end{sloppypar}

\sidecaptionvpos{figure}{c}

\subsection{Spatiotemporal System}
\begin{figure}[b]
  \centering
  \includegraphics[width=\columnwidth,keepaspectratio]{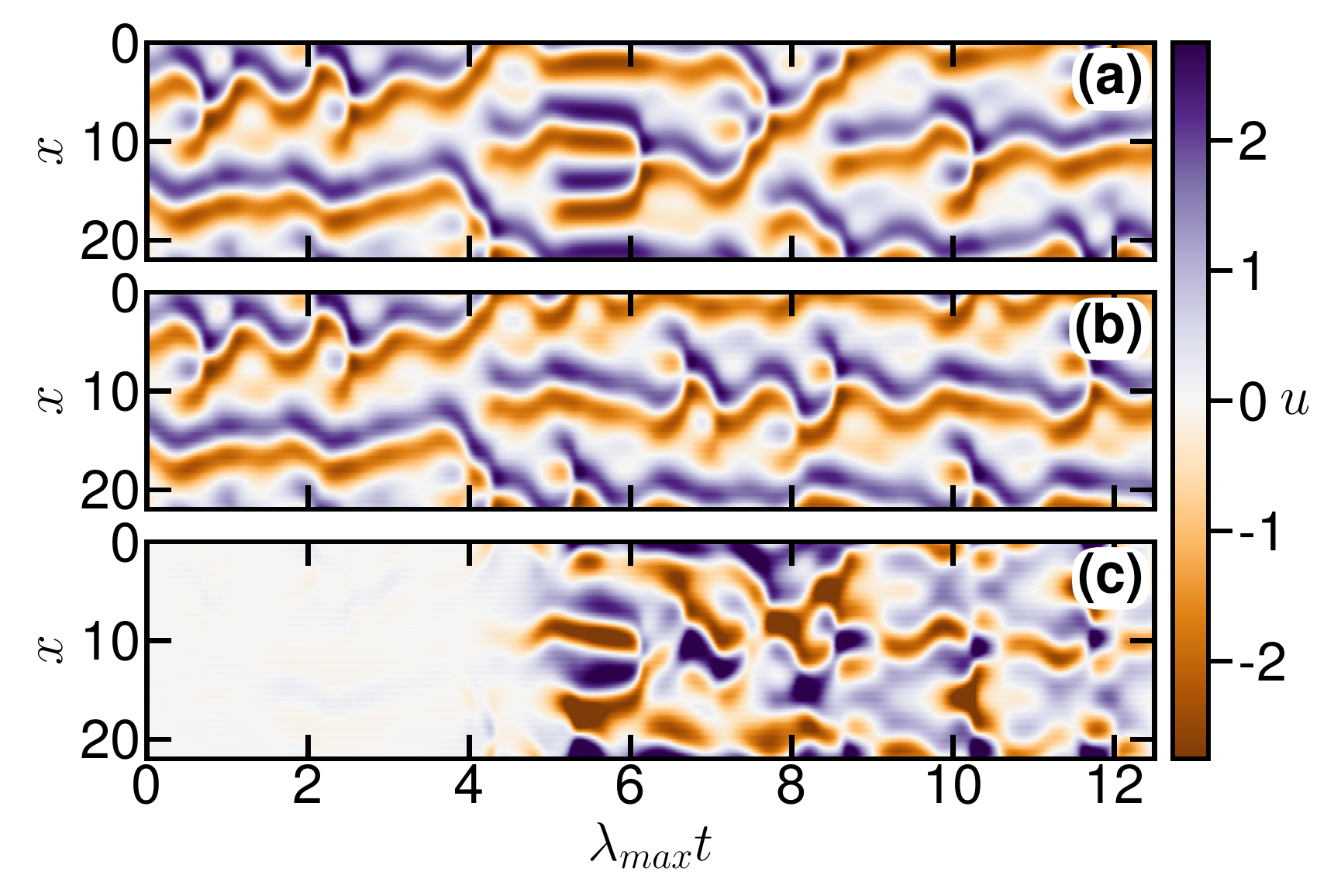}
  \caption{Prediction of Kuramoto–Sivashinsky equation with $L=22$ and $D = 64$ grid points. The $x$-axis shows Lyapunov time, where $\lambda_{max}=0.043$. Using $\xi=1$ and a quantile threshold of 0.5, TreeDOX selected $k=26$ according to Eqn.~\ref{k}. Using $\vec{FI}$ from ETR \#1, $p=585$ of $kD=1,664$ possible features are greater than $FI_0$. \textbf{(a,b)} The test and forecasted dynamics, respectively. \textbf{(c)} The difference between the test and forecasted dynamics. Here, there were 46,837 and 1,162 training and testing samples used, respectively.}
  \label{fig:ks}
\end{figure}

Next, we test TreeDOX on a chaotic spatiotemporal system by forecasting the Kuramoto–Sivashinsky (KS) system: $u_t + u_{xxxx} + u_{xx} + uu_x = 0$, where $x\in[0, L]$. To generate the training and testing data, we used length $L = 22$ and $D = 64$ grid points to discretize the domain, then used the exponential time-differencing fourth-order Runge–Kutta (ETDRK4) method to evolve the system for 50,000 iterations with $\Delta t=0.25$, random initial data, and periodic boundary conditions \cite{kassam2005fourth,pathak2018model}. After removing 2,000 transient points, we use 46,837 points for training and 1,162 points for testing. This was done to produce 12.5 Lyapunov time for the testing phase, assuming $\lambda_{max}=0.043$ \cite{edson2019lyapunov}. Using $\xi=1$ and a quantile threshold of 0.5, TreeDOX selected $k=26$ according to Eqn.~\ref{k}. Using $\vec{FI}$ from ETR \#1, $p=585$ of $kD=1,664$ possible features are greater than $FI_0$. Fig.~\ref{fig:ks} demonstrates promising results for predicting spatiotemporal time series such as the Kuramoto–Sivashinsky system, which are comparable to RC results \cite{pathak2018model}.

\sidecaptionvpos{figure}{c}
\begin{SCfigure}[50]
    \centering
    \includegraphics[width=.5\columnwidth,keepaspectratio]{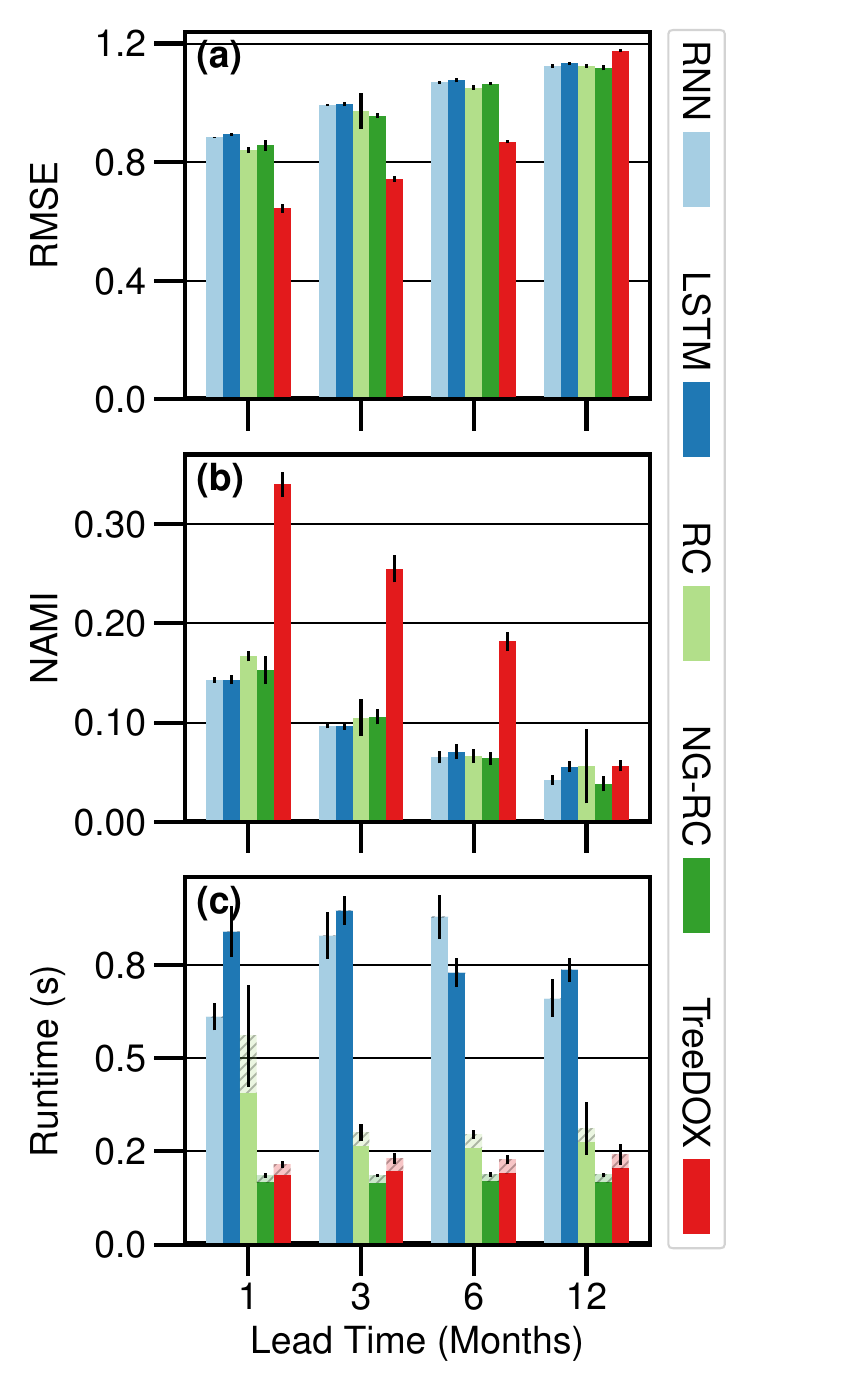}
    \caption{Summary of SOI forecasts, where lead time is 1, 3, 6, and 12 months. RNN, LSTM, RC, NG-RC, and TreeDOX are light blue, blue, light green, green, and red, respectively. Bars represent means, and error bars are $\pm 1$ standard deviation, where there are 20 realizations of each model. With $\xi=1$ and a quantile threshold of 0.5, TreeDOX selects $k=29$ in the AMI-based prescription in Eqn.~\ref{k}. \textbf{(a)} Root Mean Square Error (RMSE) between forecasted and test data. \textbf{(b)} Normalized Average Mutual Information (NAMI) between predicted and test data, where the raw AMI values are scaled by the AMI of the test data with itself. \textbf{(c)} Runtime, in seconds, of combined model training and testing, where batch predictions were used when possible. Runtime does not include the grid search for hyperparameter tuning. Note that the hatched regions stacked upon each bar display the testing runtime, while solid sections display training time. Here, there were 1,416 and 493 training and testing points used, respectively.}
    \label{fig:soi_comparison}
\end{SCfigure}

\begin{figure}[h]
    \centering
    \includegraphics[width=\columnwidth,keepaspectratio]{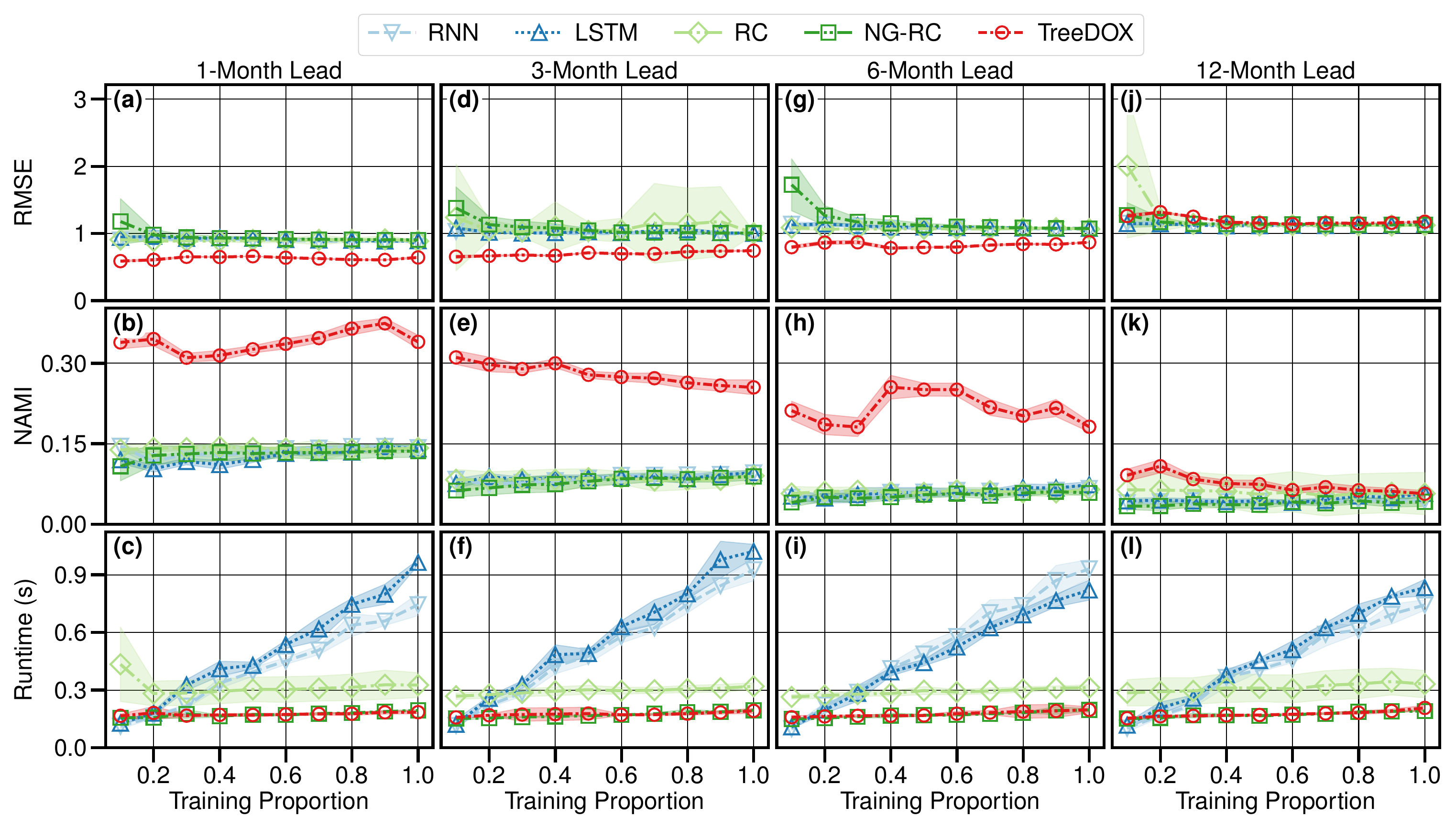}
    \caption{Open-loop prediction results for SOI data where the length of training data is varied, and the test data remains fixed. RNN, LSTM, RC, NG-RC, and TreeDOX are light blue, blue, light green, green, and red, respectively. See legend for respective line and marker styles for TreeDOX, NG-RC, RC, LSTM, and RNN results. Colored areas represent $\pm$ 1 standard deviation in results over 20 realizations. Columns 1 through 4 group results by the lead, and rows 1 and 2 display Root Mean Square Error (RMSE) and Normalized Average Mutual Information (NAMI) between predicted and test, respectively. Row 3 shows the runtime, in seconds, of training and testing (not including hyperparameter tuning).}
    \label{fig:soi_vary_training}
\end{figure}

\subsection{Real-World Dataset}
To stress test TreeDOX for a real-world chaotic time series we attempt to forecast the Southern Oscillation Index (SOI), a useful climate index with a relationship to the El Ni\~no - La Ni\~na climate phenomena, Walker circulation, drought, wave climate, and rainfall \cite{ropelewski1987extension, kiladis1988southern, trenberth1997definition, power2011impact, harisuseno2020meteorological, ranasinghe2004southern, chowdhury2010australian}. SOI is calculated as the z-score of the monthly mean sea level pressure between Tahiti and Darwin \cite{ropelewski1987extension}. SOI encodes a high dimensional chaotic system into one dimension, resulting in unpredictable data which proves difficult for current models to predict \cite{jgrd50190}.

Historically recorded SOI data features monthly values from January 1866 to July 2023 \cite{ropelewski1987extension, soi-data1}. January 1866 to December 1983 is reserved as training data, while the rest is used for testing, producing 1,416 and 493 training and testing points, respectively. See Supplementary Fig.~5 for a visualization of the SOI data. Due to the difficult nature of forecasting SOI data, TreeDOX and all other models tested are allowed to perform open-loop forecasting, meaning once a model predicts the next value, the correct test value is provided to the model before making the next prediction. To investigate the ability of TreeDOX and other models to make realistic forecasts, each model is trained to predict several months in advance (denoted as `lead' time) \cite{yan2020temporal}. Fig.~\ref{fig:soi_comparison} displays the SOI prediction results for TreeDOX and other models. See Supplementary Fig.~6 for example 1-month lead forecasts associated with Fig.~\ref{fig:soi_comparison}. Lastly, see Supplementary Fig.~7 for rudimentary k-fold predictions, demonstrating the generalizability of TreeDOX.

\section{Discussion}
\subsection{Accuracy}
One may observe that for both the H\`enon map and Lorenz systems, TreeDOX captures their respective chaotic attractors with a similar correlation dimension. Furthermore, self-evolved TreeDOX forecasts for both the Lorenz system and Kuramoto-Sivashinsky equation show similar accuracy to state-of-the-art methods, such as LSTM and NG-RC, despite its lack of hyperparameter tuning \cite{pathak2018model}. Observe in Fig.~\ref{fig:lorenz_comparison} that TreeDOX (similar to RC and NG-RC) achieves an accurate self-evolved forecast until roughly 9 Lyapunov times. Lastly, open-loop TreeDOX matches the performance of other current models in the prediction of SOI data, with comparably lower RMSE and higher AMI, supporting the ability of TreeDOX' explicit delay-overembedded memory to capture fading memory similar to that of the implicit memory of LSTM, NG-RC, and other models.

\subsection{Speed and Complexity}
While TreeDOX training cannot outperform the linear training scheme of RC and NG-RC, it benefits from computational simplicity and the usage of classical, well-understood regression tree ensemble methods. Additionally, we emphasize that the suggested hyperparameter prescription scheme allows TreeDOX to skip the time-consuming and subjective hyperparameter tuning process involved in other current methods, meaning that in practice, TreeDOX may be faster and more hassle-free than the alternatives. Due to the ensemble nature of ETRs and RFs, each individual regression tree is entirely independent, meaning the potential for GPU-parallelized training exists. Under the \texttt{RAPIDS} ecosystem, \texttt{cuML}'s \texttt{RandomForestRegressor} implements RFs with GPU-parallelization, which provides substantial speedups compared to \texttt{Scikit-learn}'s \texttt{RandomForestRegressor} or \texttt{ExtraTreesRegressor} for users matching the software and hardware requirements \cite{RAPIDS, raschka2020machine, scikit-learn}. Furthermore, the recently developed \texttt{Hummingbird} package allows for trained \texttt{Scikit-learn} tree ensembles to be converted into a tensor equivalent for tree traversal, allowing for very fast model prediction using GPU resources \cite{nakandalam2020taming}.

One weakness of TreeDOX lies in the poor computational complexity in regards to the dimensionality of data (and very large training length). Decision trees scale in computational complexity not only with the depth (and thus complexity of data), but also with the number of features and training samples. Since the number of features scales with the dimension of training data, it follows that the computational complexity also scales with the dimension. It is not straightforward to express the computational complexity of decision trees due to their non-fixed training scheme, but it is commonly thought that the complexity scales linearly with depth and the number of features, and super-linearly with the number of samples (usually between $O(n\log{n})$ and $O(n^2\log{n})$ where $n$ is the number of samples). Therefore the computational complexity of an ensemble of decision trees is likely between $O(N_{trees}\cdot{depth}\cdot kDn\log{n})$ and $O(N_{trees}\cdot{depth}\cdot kDn^2\log{n})$ where $N_{trees}$ is the number of trees in the ensemble, $depth$ is the average depth of trees, $k$ is the delay overembedding dimension, $D$ is the dimension of training data, and $n$ is the number of samples. However, one may improve runtime for TreeDOX on high-dimension data (or data with very long training length) by limiting the number of trees in the ensemble and the maximum allowable tree depth.

\sidecaptionvpos{figure}{c}
\begin{SCfigure}[50]
    \centering
    \includegraphics[width=.5\columnwidth,keepaspectratio]{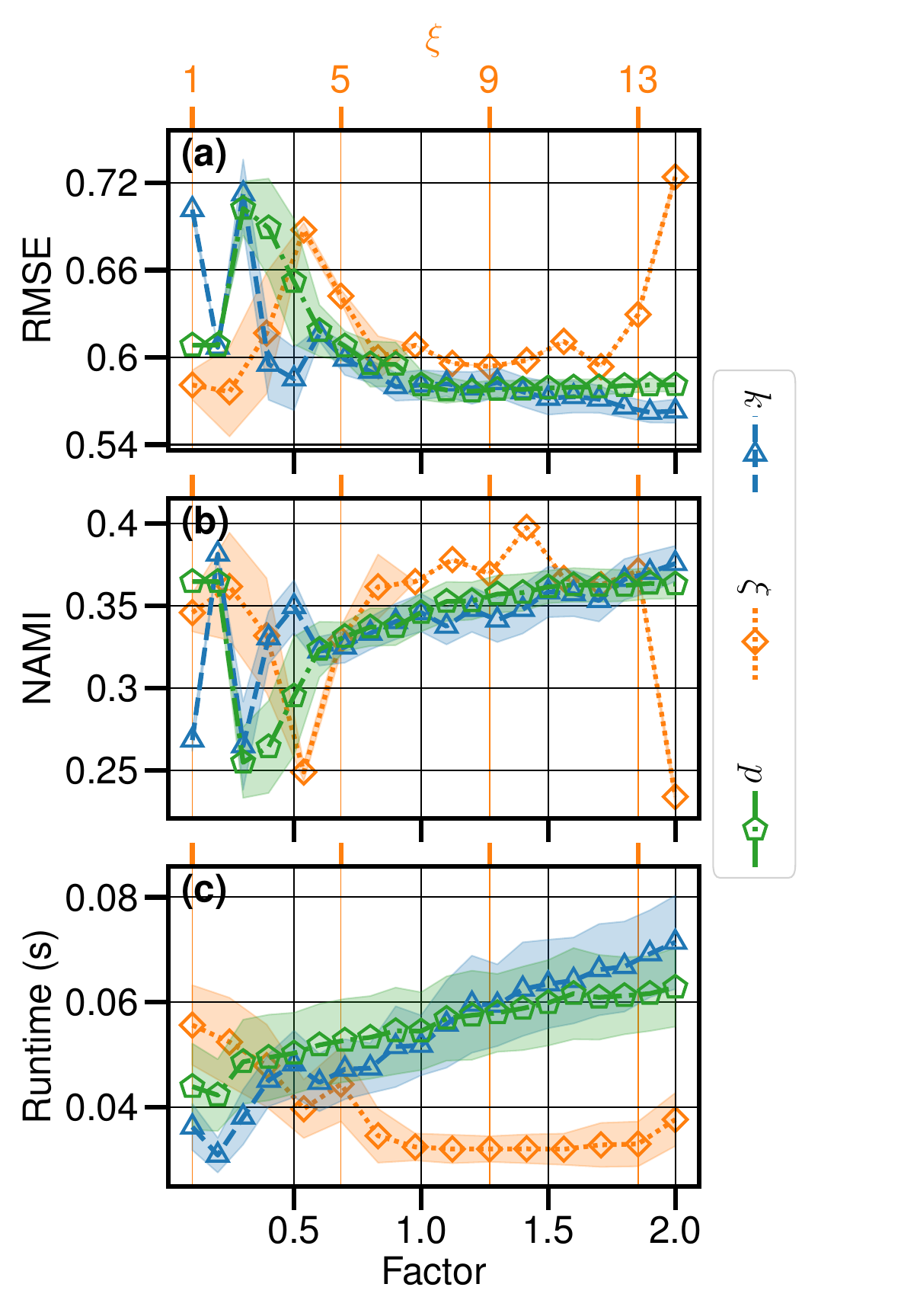}
    \caption{Sensitivity to hyperparameters in SOI training, where blue triangles are the overembedding dimension, $k$, orange squares are the successive time lags, $\xi$, and green pentagons are the number of reduced features, $p$. Shaded regions indicate $\pm$1 standard deviation with 1000 realizations of each point. Training and test data is equivalent to Fig.~\ref{fig:soi_comparison}. \textbf{(a)} Root Mean Square Error (RMSE) and \textbf{(b)} Normalized Average Mutual Information between test and predicted SOI data. \textbf{(c)} Runtime of combined model training and testing in seconds. `Factor' is the multiplicative factor applied to each hyperparameter, meaning 1.0 is simply the prescribed value of the given hyperparameter. The orange twin axis shows the value of $\xi$ independently since $\xi$ must be a natural number. Here, we take the default values of $k$ and $p$ as described earlier with an associated quantile threshold of 0.5, and the default value of $\xi$ is 1.}
    \label{fig:soi_sensitivity}
\end{SCfigure}

\subsection{Remark on Training Data Length}
Aside from eliminating hyperparameter tuning, we use SOI data (with a lead value of 1) to demonstrate that TreeDOX does not require as much training data as the alternative methods. We keep the test data equivalent to that used in Fig.~\ref{fig:soi_comparison} and vary the length of the training data, where the training region kept a temporally contiguous stretch terminating right before the test data. Fig.~\ref{fig:soi_vary_training} displays the results of this experiment. Note that the RMSE and NAMI of TreeDOX remain mostly stable despite the length of the training data. Furthermore, the RMSE is usually lower, and NAMI is usually higher than the compared methods, indicating the superior accuracy of TreeDOX. The bottom row of Fig.~\ref{fig:soi_vary_training} shows that, despite TreeDOX not beating the speed of linear RC and NG-RC training, it scales better than RNN and LSTM as the length of training data increases. We reiterate that these runtime results do not report any hyperparameter tuning required for RC, NG-RC, RNN, and LSTM, and therefore, TreeDOX may be faster to use in practice.

\subsection{Robustness to Hyperparameter Prescriptions}
We demonstrate the resilience of TreeDOX to the value of our prescribed hyperparameters by testing the sensitivity of TreeDOX predictions of SOI data to changes in the delay overembedding dimension, $k$, the successive time lag, $\xi$, and the number of reduced features, $p$. It is important to note that here we are not providing a proof of the low sensitivity of TreeDOX to these hyperparameters, but rather a heuristic numerical experiment with SOI data to demonstrate with a real-world scenario, that hyperparameter tuning is not necessary. Fig.~\ref{fig:soi_sensitivity} shows the RMSE, NAMI, and runtime results when individually varying the above hyperparameters (while keeping the others fixed to their respective prescribed values) and predicting SOI data. Note that a `factor' value of 1.0 indicates the respective hyperparameter is equivalent to its prescribed value, where the default values of $k$ and $p$ are prescribed as expressed earlier in the manuscript, and the default value for $\xi$ is 1 as described earlier. Let us first consider $k$ and $p$. Intuitively, increasing the value of any of these hyperparameters will increase model accuracy at the expense of training complexity, which scales linearly with all three hyperparameters, except for $p$, which will eventually cause the runtime to reach a plateau when $p=kD$. Fig.~\ref{fig:soi_sensitivity}\textbf{(a)} and Fig.~\ref{fig:soi_sensitivity}\textbf{(b)} show that as the respective hyperparameter is increased, model accuracy plateaus while the runtime in Fig.~\ref{fig:soi_sensitivity}\textbf{{c}} does, in fact, increase roughly linearly. Observe the lack of the aforementioned plateau in the runtime curve as $p$ varies; this is likely due to the maximum value of $p$ in this experiment not reaching its maximum value $kD$. Note that the prescribed hyperparameter values manage to balance runtime while lying on the accuracy plateau. Next, we investigate $\xi$, in which one might expect an increase in $\xi$ to impair model accuracy due to dropping short-term delay states while lowering runtime due to the respective decrease in k according to Eqn.~\ref{k}. Fig.~\ref{fig:soi_sensitivity} supports this hypothesis while also demonstrating $\xi$'s plateau effect on runtime, likely due to $\xi$ being in the denominator of Eqn.~\ref{k} and the computation complexity of training scaling linearly with $k$. Of course, we recommend selecting the smallest such possible value of $\xi$ that is computationally reasonable due to its inverse effect on accuracy.

\subsection{Final Remarks}
With the increasing availability of data and computational power to analyze it, the need for effective and user-friendly time series forecasting methods is on the rise. Existing state-of-the-art methods, such as RC and LSTM, offer a powerful ability to meet the need for time series forecasting, but can be difficult to use in practice due to their sensitivity to and required tuning of hyperparameters. We propose an alternative in the form of a time delay overembedded and Extra Tree Regressor-based algorithm for autonomous feature selection and forecasting, which does not require hyperparameter tuning.

While the development of TreeDOX was focused on ease-of-use rather than on surpassing the accuracy of modern forecasting models, after testing it on a variety of prototypical discrete, continuous, and spatiotemporal systems, we find that TreeDOX provides comparable or better performance to current methods such as RC and LSTM. We also demonstrate the efficacy of TreeDOX to predict realistic data with SOI open-loop forecasts and again discover TreeDOX's similar performance to LSTM, NG-RC, and other state-of-the-art forecasting models.


\section{Acknowledgments}
We would like to acknowledge Research Computing at the Rochester Institute of Technology for supplying computational resources during the course of this work \cite{https://doi.org/10.34788/0s3g-qd15}. The research of E.B. is supported by the ONR, ARO, DARPA RSDN and the NIH and NSF CRCNS. N.M. is supported by the National Science Foundation (NSF), Award Number: 2434716. Any opinions, findings, and conclusions or recommendations expressed in this material are those of the author(s) and do not necessarily reflect the views of the NSF.  

\section{Data and Code Availability}
Data and code are available on our Github repository: \url{https://github.com/amg2889/TreeDOX_Tree-based_Learning_for_High-Fidelity_Prediction_of_Chaos}

\section{Author Contributions}
A.G., K.R., and N.M. designed the research. A.G. carried out the research. A.G. wrote the first draft of the manuscript. A.G., K.R., E.B., and N.M. reviewed and edited the manuscript. E.B. and N.M. supervised the research.

\section{Competing Interests}
The authors declare no competing interests.



%


\clearpage

\pagebreak
\setcounter{equation}{0}
\setcounter{figure}{0}
\setcounter{table}{0}
\setcounter{page}{1}
\setcounter{section}{0}
\makeatletter
\renewcommand{\figurename}{Supplementary Fig.}
\renewcommand{\tablename}{Supplementary Table}
\renewcommand{\bibnumfmt}[1]{[S#1]}
\renewcommand{\citenumfont}[1]{S#1}

\title{Supplementary Information: Tree-based Learning for High-Fidelity Prediction of Chaos}
\maketitle

\newcommand{\ccbest}{\cellcolor[HTML]{cccccc}}


\section{Introduction}
This supplemental information provides additional visualizations and results on TreeDOX training, forecasting, and data used for testing.

\section{Method}
\begin{figure}
    \centering
    \includegraphics[width=\columnwidth,keepaspectratio]{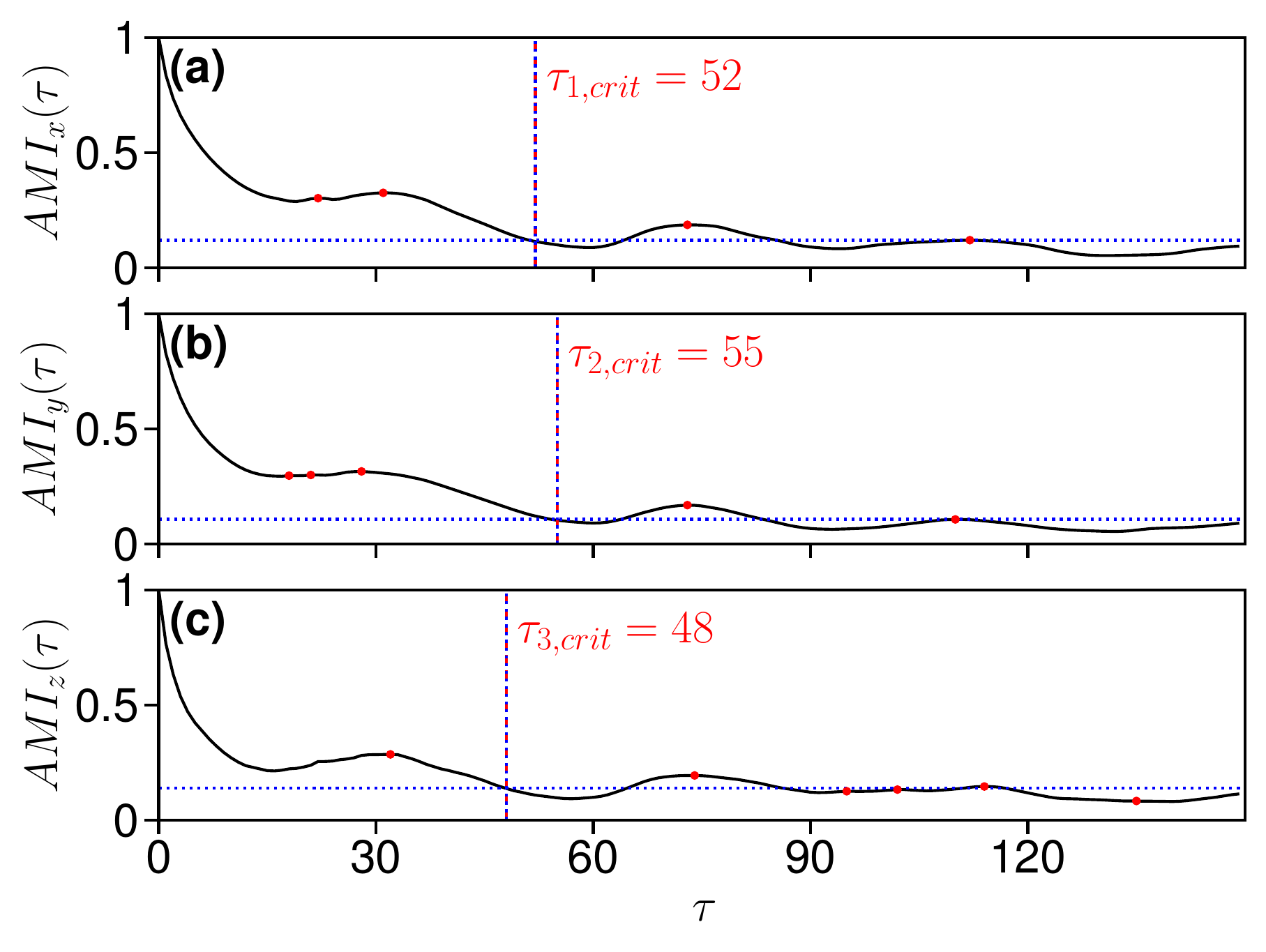}
    \caption{Average mutual information (AMI) of Lorenz System training data where $dt=0.01$. $AMI_{cdot}(\tau)$ indicates the AMI of the respective coordinate with its copy delayed by $\tau$ states, where $x$, $y$, $z$ coordinates are shown in \textbf{(a,b,c)}, respectively. Red dots indicate the local maxima of the AMI curves. Dotted blue lines indicate the measured 0.5th quantile of the respective AMI curve (horizontal) and the time lag $\tau$ at which the AMI curve drops below the given quantile. The vertical red dashed line indicates the selected $\tau_{i,crit}$. Since the crossing of the 0.5th quantile occurs after the first local minima, $\tau_{i,crit}$ is taken to be the $\tau$ for which the 0.5th quantile is crossed. Otherwise, $\tau_{i,crit}$ would be taken as the $\tau$ for which the first local maxima of the respective AMI curve occurs. See the main manuscript for more details.}
    \label{fig:ami}

\end{figure}

To prescribe a value for $k$ (delay overembedding dimension), average mutual information (AMI) is applied to each dimension of training data, and with a given p-value selected, a time delay $\tau_{crit}$ is found where the respective AMI crosses under the p-value. See the main manuscript for more information about how $k$ is prescribed from the $\tau_{crit}$ values. Supplementary Fig.~\ref{fig:ami} shows the AMI calculation for Lorenz System training data where $dt=0.01$. One may observe the sharp drop in mutual information as the time delay $\tau$ increases, indicating that the lags are losing information about the current state of the system. $\xi$ (the successive time delays in the time delay overembedding) should be selected to the smallest value possible such that the resulting time and space complexity of TreeDOX training is not too great a burden.

\begin{figure}
    \centering
    \includegraphics[width=\columnwidth,keepaspectratio]{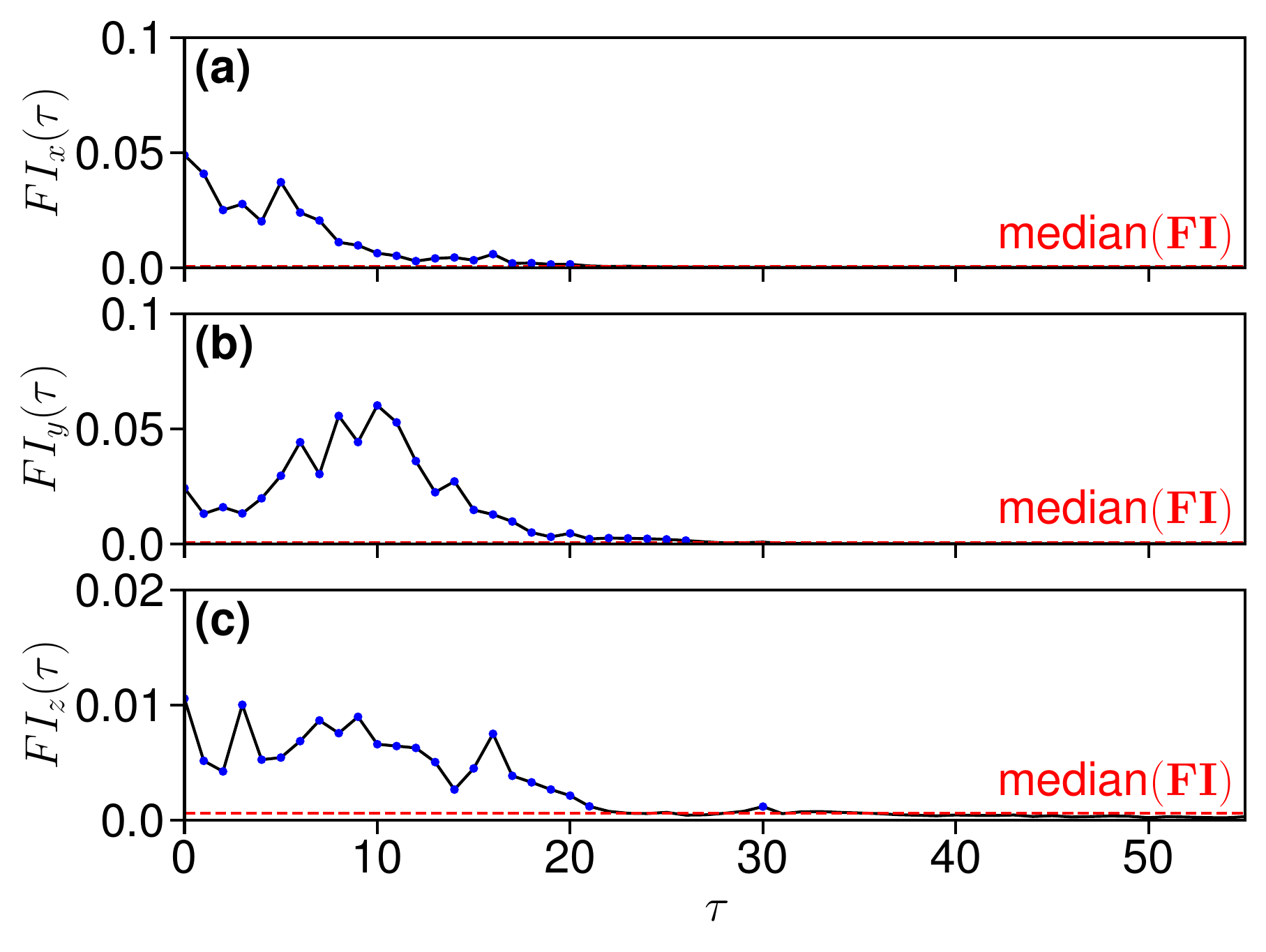}
    \caption{Gini feature importances measured by ETR \#1 in the main manuscript for Lorenz System training data where $dt=0.01$. For clarity, $FI$ is unflattened to show the respective lags chosen for each of the three dimensions, where $x$, $y$, and $z$ are shown in \textbf{(a,b,c)}, respectively. Blue dots indicate features that are greater than $FI_0$, which is the minimum feature importance that passes a 5\% t-test that the population mean of the respective feature importance value is greater than the median of all $\vec{FI}$. Note that, especially in \textbf{(c)}, one can observe that there are features that are, despite their nominal feature importance being greater than the median, not selected due to their variance indicating a lack of statistical significance when compared to the null hypothesis that the population mean is less than or equal to the median.}
    \label{fig:FIs}
\end{figure}

The number of features to use in final training, $p$, is prescribed by comparing the Gini feature importances, $FI$, from ETR \#1 in the main manuscript to a null rate, $FI_0$. As described in the main manuscript, since ensemble tree methods contain an array of feature importances for each respective tree, we compute a t-test on the realizations of each feature's importance across all trees. To start, the null hypothesis is $H_0:~\mu \leq 0$, and the alternative hypothesis is $H_1:~\mu>0$, where $\mu$ is the population mean of the feature importance. We set $FI_0$ as the minimum nominal feature importance whose observations pass the t-test with a significance threshold of 5\%, then take $p$ as the number of features whose feature importance is greater than or equal to $FI_0$. The reasoning here is that Gini importance typically has high variance; however, one should not discount features whose Gini importance has both a high mean and variance, which may cause the respective feature not to pass the t-test with high significance. Lastly, for scenarios where many time lags are considered or the given training data has a high dimension, the user may be unsatisfied if the t-test results in a majority of features making it through the reduction process, leading to high computational complexity when doing the final training and testing of ETR\#2. Therefore, we suggest that if the initial t-test selects $p>kD/2$ (meaning more than half the possible features remain after reduction), then one should repeat the t-test with $H_0:~\mu\leq\text{median}(\vec{FI})$ and $H_1:~\mu>\text{median}(\vec{FI})$. Supplementary Fig.~\ref{fig:FIs} displays the measured $\vec{FI}$ for Lorenz System training data where $dt=0.01$. Observe, like Supplementary Fig.~\ref{fig:ami}, the sharp drop-off of $\vec{FI}$ as the time delay increases. While this indicates similar phenomena, instead of being interpreted as a drop in the amount of mutual information between a delayed state and the current state, it may be interpreted by ETR \#1 as finding the delayed state to have an insignificant ability to make predictions of the current state. While delay states with the respective feature importance under $FI_0$ may still be considered useful for prediction, their removal provides two major advantages: reducing the time and space complexity of TreeDOX training and enhancing the model's generalizability. Due to the training speedup after feature reduction, one may reasonably increase the number of estimators in the ETR ensemble to further improve results with little reduction in training performance.

\section{Logistic Map}
To further reinforce the ability of TreeDOX to predict discrete chaotic systems, we apply the model to the logistic map: $x_{n+1} = rx_n(1-x_n)$ where $x\in[0,1]$ and $r\in[0,4]$. To start, we selected a variety of values for the parameter $r$ - namely 0.5, 2, 3.2, 3.5, 3.56, 3.6, 3.7, 3.8, and 3.9 - and trained TreeDOX with 10,000 points generated from the iteration scheme with $x_0 = 0.5$ and 10 transient points removed. Then, we test on the next 1,000 points. Using $\xi=1$, $k$ is prescribed based on Eqn.~1 in the main manuscript. Likewise, $p$ is prescribed using the t-test discussed in the main manuscript. See Supplementary Table~\ref{tab:logistic_hyperparams} for respective values. We plot $x_{n+1}$ against $x_n$ to determine if our forecasted data lies on the above parabola of the true dynamics, confirming TreeDOX has learned the underlying system. The results seen in Supplementary Fig.~\ref{fig:lmdynamics} confirm the algorithm's preservation of the generating dynamics of the underlying system since the forecasted data does, in fact, lie on the parabola $x_{n+1} = rx_n(1-x_n)$.

\begin{table}[h]
    \centering
    \begin{tabular}{|c||c|c|c|c|c|c|c|c|c|}
        \hline
        ~$r$~ & ~0.5~ & ~2~ & ~3.2~ & ~3.5~ & ~3.56~ & ~3.6~ & ~3.7~ & ~3.8~ & ~3.9~ \\
        \hhline{|=||=|=|=|=|=|=|=|=|=|}
        ~$k$~ & 2 & 2 & 5 & 5 & 9 & 18 & 20 & 21 & 18 \\
        \hline
        ~$p$~ & 1 & 2 & 2 & 2 & 3 & 7 & 10 & 9 & 7 \\
        \hline
    \end{tabular}
    \caption{Hyperparameter prescriptions associated with Supplementary Fig.~\ref{fig:lmdynamics}.}
    \label{tab:logistic_hyperparams}
\end{table}

Next, we attempt to use a much larger set of training data and compare the last point in our forecasted results to the true bifurcation diagram of the logistic map. For each randomly generated value of $r\in[2.75,4]$, we train our algorithm with a length-2,000 time series of the true logistic evolution with a randomly generated $x_0\in[0,1]$ (after removing 25 transient points). We use $\xi=1$ and prescribe $k$ and $p$ according to the hyperparameter prescription described earlier. Lastly, we evolve our model for 1,000 points and plot the last value in the forecasted time series. Supplementary Fig.~\ref{fig:lmbifurcation} confirms that our forecasting method captures the logistic dynamics well enough to recreate the bifurcation diagram with a substantial resolution.

\sidecaptionvpos{figure}{c}
\begin{SCfigure}[30]
    \centering
    \includegraphics[width=0.7\columnwidth,keepaspectratio]{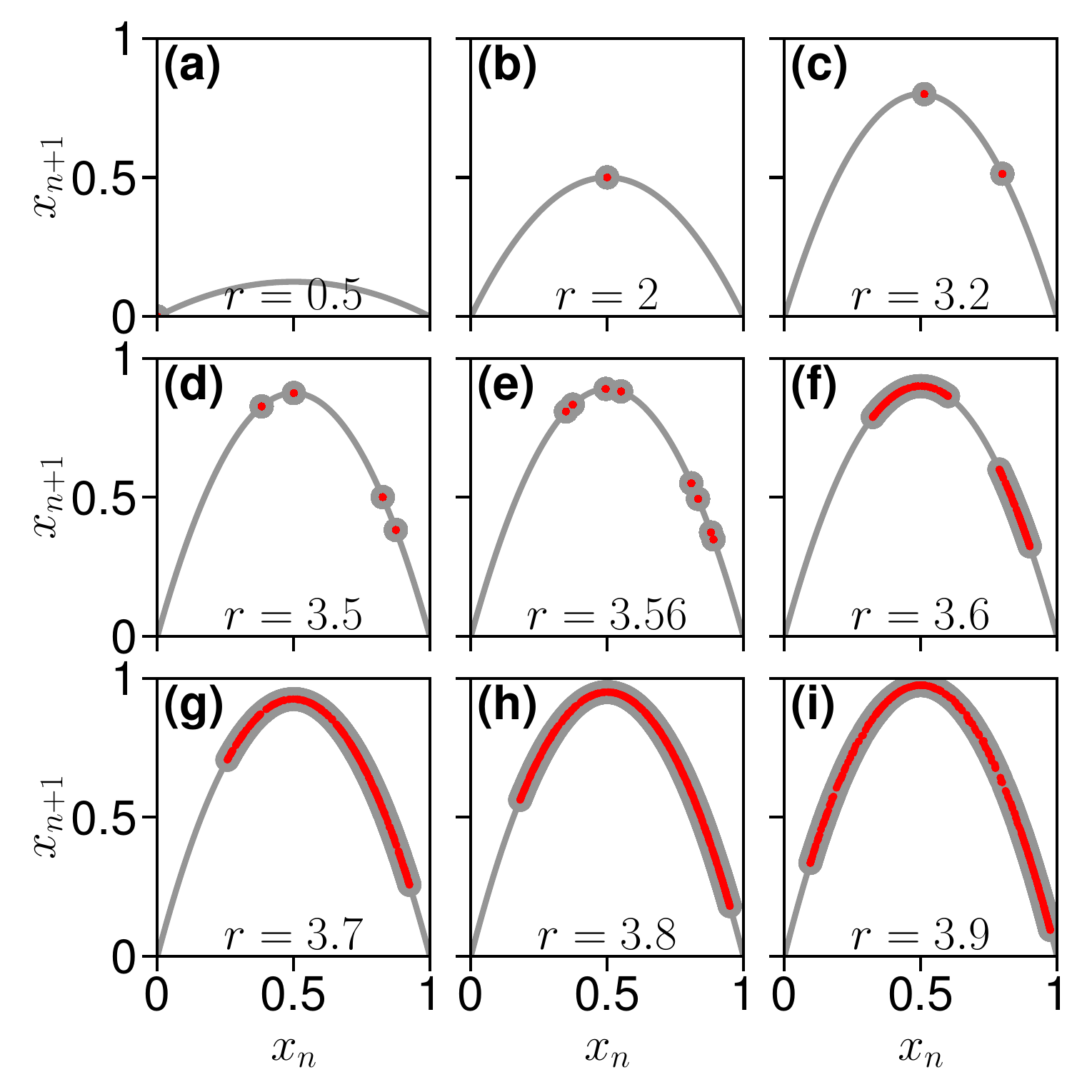}
    \caption{Recreation of logistic map dynamics for a range of $r$ values, where the gray parabola is the true dynamics ($x_{n+1} = rx_n(1-x_n)$) and the forecasted and test data are red and gray, respectively. \textbf{(a,b,c,d,e,f,g,h,i)} Test and predicted dynamics for $r$ values of 0.5, 2, 3.2, 3.5, 3.56, 3.6, 3.7, 3.8, and 3.9, respectively. Here we used an initial value of $x_0=0.5$, and 10,000 training and 1,000 test points, respectively.}
    \label{fig:lmdynamics}
\end{SCfigure}

\begin{figure}
    \includegraphics[width=\columnwidth,keepaspectratio]{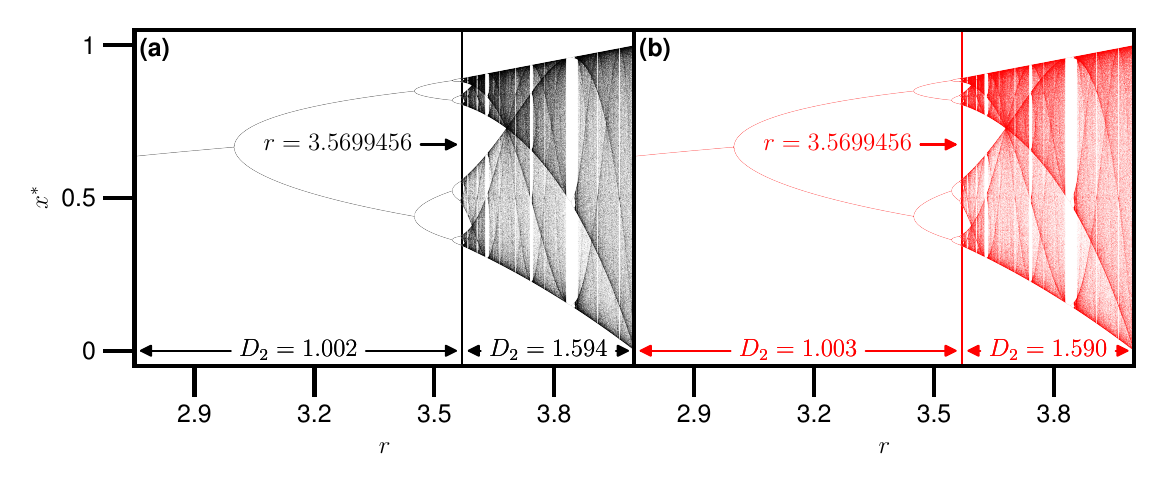}
    \caption{Learning the bifurcation diagram of the logistic map, where the forecasted and test data are red and black, respectively. $x^*$ is the last point in the respective time series, and $r$ is the logistic parameter. \textbf{(a,b)} Test and predicted bifurcation diagrams, respectively. $D_2$ is the correlation dimension calculated by the region of data specified by the respective arrows. Here, we used a random initial value $x_0\in (0,1)$, and 2,000 training and 1,000 testing points, respectively.}
    \label{fig:lmbifurcation}
\end{figure}

Lastly, we perform a numerical experiment over a varying number of trajectories and training length to investigate the efficacy of TreeDOX with as little bias as possible. We forecast on 20 sets of data and show the resulting summary statistics in Supplementary Table~\ref{tab:logistic}. Each set of data is generated by varying the initial conditions: $x_0=0.25+0.025m$, where $0\leq m \leq 19$ is the index of the dataset. Then, 25,000 and 100 training and testing points are created by evolving the logistic map. When the training length is varied (as in Supplementary Table~\ref{tab:logistic}), the actual training and testing data are fixed, but each model is fed the right-most time window (with the specified training length) of the total 25,000 training points.

\renewcommand{\arraystretch}{1.25}
\begin{table}
    \centering
    \begin{tabular}{C{4em}|c||c|c|c|c|c|}
        \hline
        Training Length & Metric & RNN  & LSTM & RC & NGRC & TreeDOX\\
        \hline\hline
        \multirow{6}{*}{2,500} 
            & RMSE &  $\bm{0.234 \pm 0.049}$  &  $0.250 \pm 0.062$  &  $0.429 \pm 0.724$  &  $0.265 \pm 0.028$  &  $0.254 \pm 0.027$   \\\cline{2-7}
            & NAMI &  $0.063 \pm 0.034$  &  $0.091 \pm 0.058$  &  $0.173 \pm 0.038$  &  $0.206 \pm 0.045$  &  $\bm{0.244 \pm 0.046}$   \\\cline{2-7}
            & Forecast Horizon &  $0.5 \pm 1.2$  &  $0.9 \pm 3.3$  &  $0.3 \pm 0.6$  &  $12.7 \pm 4.4$  &  $\bm{27.8 \pm 4.6}$   \\\cline{2-7}
            & Train+Test Time (s) &  $3.658 \pm 1.063$  &  $3.403 \pm 1.911$  &  $0.491 \pm 0.277$  &  $\bm{0.282 \pm 0.091}$  &  $0.631 \pm 0.018$   \\\cline{2-7}
            & Tune Time (s) &  $42.107 \pm 1.037$  &  $50.584 \pm 1.888$  &  $12.628 \pm 1.012$  &  $\bm{6.627 \pm 0.242}$   & \\\cline{2-7}
            & Total Time (s) &  $45.764 \pm 1.589$  &  $53.987 \pm 3.204$  &  $13.118 \pm 0.983$  &  $6.909 \pm 0.243$  &  $\bm{0.631 \pm 0.018}$   \\\cline{2-7}
        \hline\hline
        \multirow{6}{*}{10,000} 
            & RMSE &  $0.290 \pm 0.057$  &  $0.286 \pm 0.055$  &  $2.047 \pm 7.771$  &  $0.267 \pm 0.039$  &  $\bm{0.230 \pm 0.025}$   \\\cline{2-7}
            & NAMI &  $0.170 \pm 0.069$  &  $0.157 \pm 0.099$  &  $0.156 \pm 0.027$  &  $0.235 \pm 0.054$  &  $\bm{0.290 \pm 0.053}$   \\\cline{2-7}
            & Forecast Horizon &  $0.2 \pm 0.8$  &  $0.1 \pm 0.4$  &  $1.4 \pm 1.4$  &  $17.8 \pm 7.9$  &  $\bm{31.6 \pm 5.8}$   \\\cline{2-7}
            & Train+Test Time (s) &  $6.638 \pm 0.973$  &  $6.582 \pm 2.615$  &  $1.037 \pm 0.350$  &  $\bm{0.515 \pm 0.061}$  &  $0.708 \pm 0.010$   \\\cline{2-7}
            & Tune Time (s) &  $93.450 \pm 1.511$  &  $122.686 \pm 5.459$  &  $27.806 \pm 2.549$  &  $\bm{13.621 \pm 1.364}$   & \\\cline{2-7}
            & Total Time (s) &  $100.087 \pm 1.995$  &  $129.267 \pm 6.055$  &  $28.843 \pm 2.625$  &  $14.136 \pm 1.359$  &  $\bm{0.708 \pm 0.010}$   \\\cline{2-7}
        \hline\hline
        \multirow{6}{*}{17,500} 
            & RMSE &  $0.281 \pm 0.052$  &  $0.278 \pm 0.060$  &  $0.411 \pm 0.531$  &  $0.254 \pm 0.034$  &  $\bm{0.251 \pm 0.030}$   \\\cline{2-7}
            & NAMI &  $0.149 \pm 0.085$  &  $0.132 \pm 0.071$  &  $0.169 \pm 0.057$  &  $0.242 \pm 0.047$  &  $\bm{0.254 \pm 0.040}$   \\\cline{2-7}
            & Forecast Horizon &  $0.1 \pm 0.7$  &  $0.1 \pm 0.7$  &  $1.0 \pm 1.1$  &  $17.4 \pm 4.7$  &  $\bm{30.1 \pm 3.6}$   \\\cline{2-7}
            & Train+Test Time (s) &  $7.387 \pm 2.699$  &  $7.590 \pm 4.713$  &  $1.300 \pm 0.168$  &  $\bm{0.800 \pm 0.074}$  &  $0.868 \pm 0.127$   \\\cline{2-7}
            & Tune Time (s) &  $143.509 \pm 2.263$  &  $194.948 \pm 6.561$  &  $44.305 \pm 3.651$  &  $\bm{19.831 \pm 1.220}$   & \\\cline{2-7}
            & Total Time (s) &  $150.896 \pm 3.529$  &  $202.538 \pm 8.182$  &  $45.604 \pm 3.648$  &  $20.630 \pm 1.232$  &  $\bm{0.868 \pm 0.127}$   \\\cline{2-7}
        \hline\hline
        \multirow{6}{*}{25,000}
            & RMSE &  $0.299 \pm 0.050$  &  $0.271 \pm 0.061$  &  $1.901 \pm 6.598$  &  $0.269 \pm 0.031$  &  $\bm{0.249 \pm 0.030}$   \\\cline{2-7}
            & NAMI &  $0.164 \pm 0.073$  &  $0.157 \pm 0.092$  &  $0.172 \pm 0.034$  &  $0.237 \pm 0.038$  &  $\bm{0.292 \pm 0.065}$   \\\cline{2-7}
            & Forecast Horizon &  $0.1 \pm 0.7$  &  $0.1 \pm 0.7$  &  $1.1 \pm 1.9$  &  $18.9 \pm 3.6$  &  $\bm{33.5 \pm 4.8}$   \\\cline{2-7}
            & Train+Test Time (s) &  $7.936 \pm 4.163$  &  $13.966 \pm 10.046$  &  $1.794 \pm 0.295$  &  $1.040 \pm 0.073$  &  $\bm{0.961 \pm 0.047}$   \\\cline{2-7}
            & Tune Time (s) &  $192.951 \pm 3.030$  &  $269.803 \pm 8.659$  &  $57.527 \pm 4.246$  &  $\bm{26.836 \pm 1.432}$   & \\\cline{2-7}
            & Total Time (s) &  $200.887 \pm 5.536$  &  $283.769 \pm 12.869$  &  $59.321 \pm 4.258$  &  $27.876 \pm 1.431$  &  $\bm{0.961 \pm 0.047}$   \\\cline{2-7}
        \hline
    \end{tabular}
    \caption{Numerical experiment with varying training length and 20 total logistic map independent trajectories. Each cell displays the mean metric across the 20 realizations, plus or minus their standard deviations. The bold cells represent the best model in the respective category. RMSE is Root Mean Square Error, NAMI is Normalized Average Mutual Information, and Forecast Horizon represents the number of time points for which the absolute residual error for all dimensions is less than the standard deviation of the training data for that respective coordinate. The forecast horizon represents the length of time for which the forecasts are qualitatively converged to the test data. Hyperparameter tuning for RNN, LSTM, RC, and NGRC uses TPE optimization (see the main manuscript) with 25 samples.}
    \label{tab:logistic}
\end{table}

\section{Southern Oscillation Index}
\begin{figure}
    \centering
    \includegraphics[width=\linewidth,keepaspectratio]{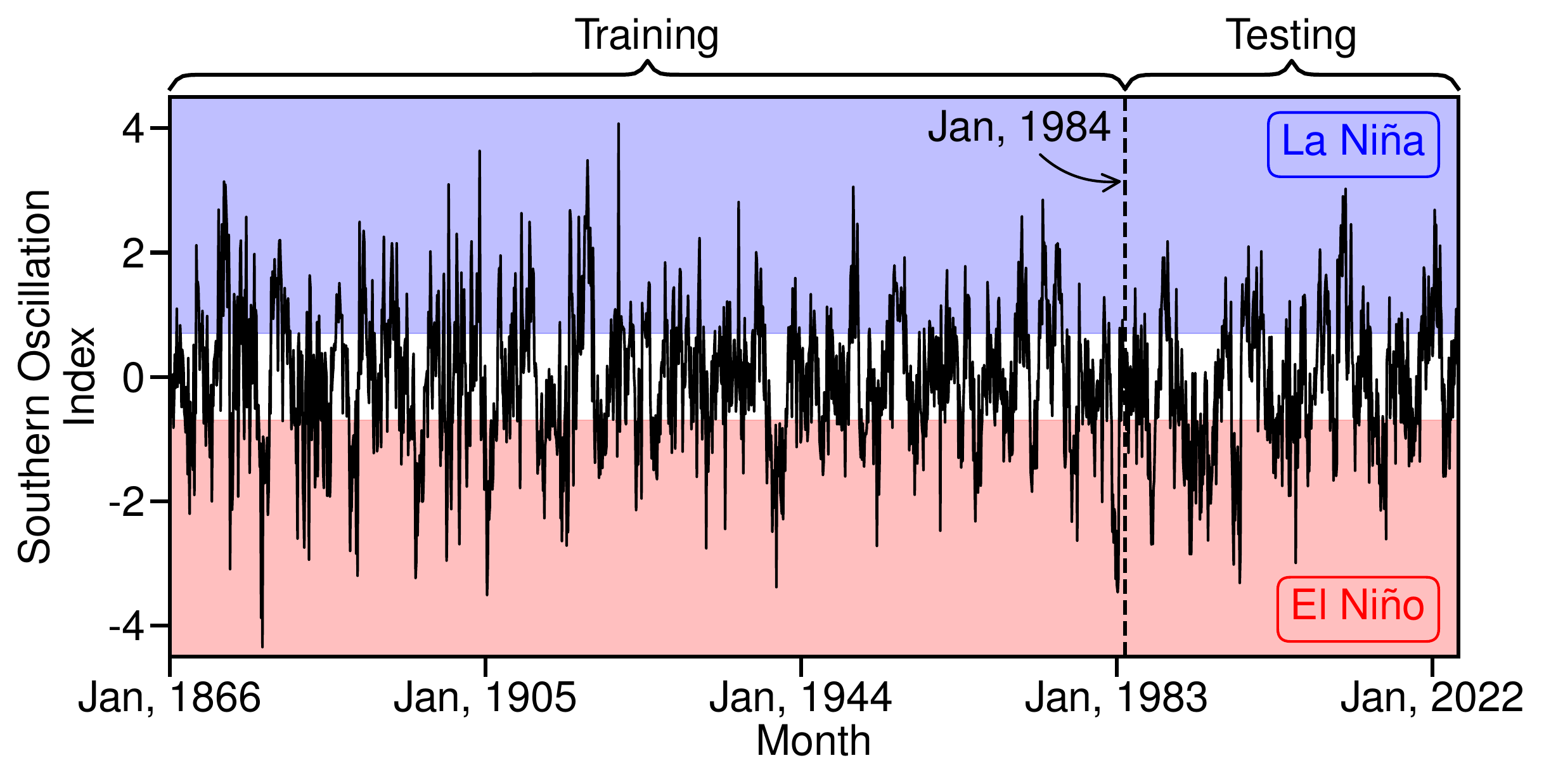}
    \caption{Historical monthly Southern Oscillation Index data, where data prior to and following January 1984 is used for training and testing, respectively, of TreeDOX and compared models.}
    \label{fig:soi_raw}
\end{figure}

\begin{figure}
    \centering
    \includegraphics[width=\linewidth,keepaspectratio]{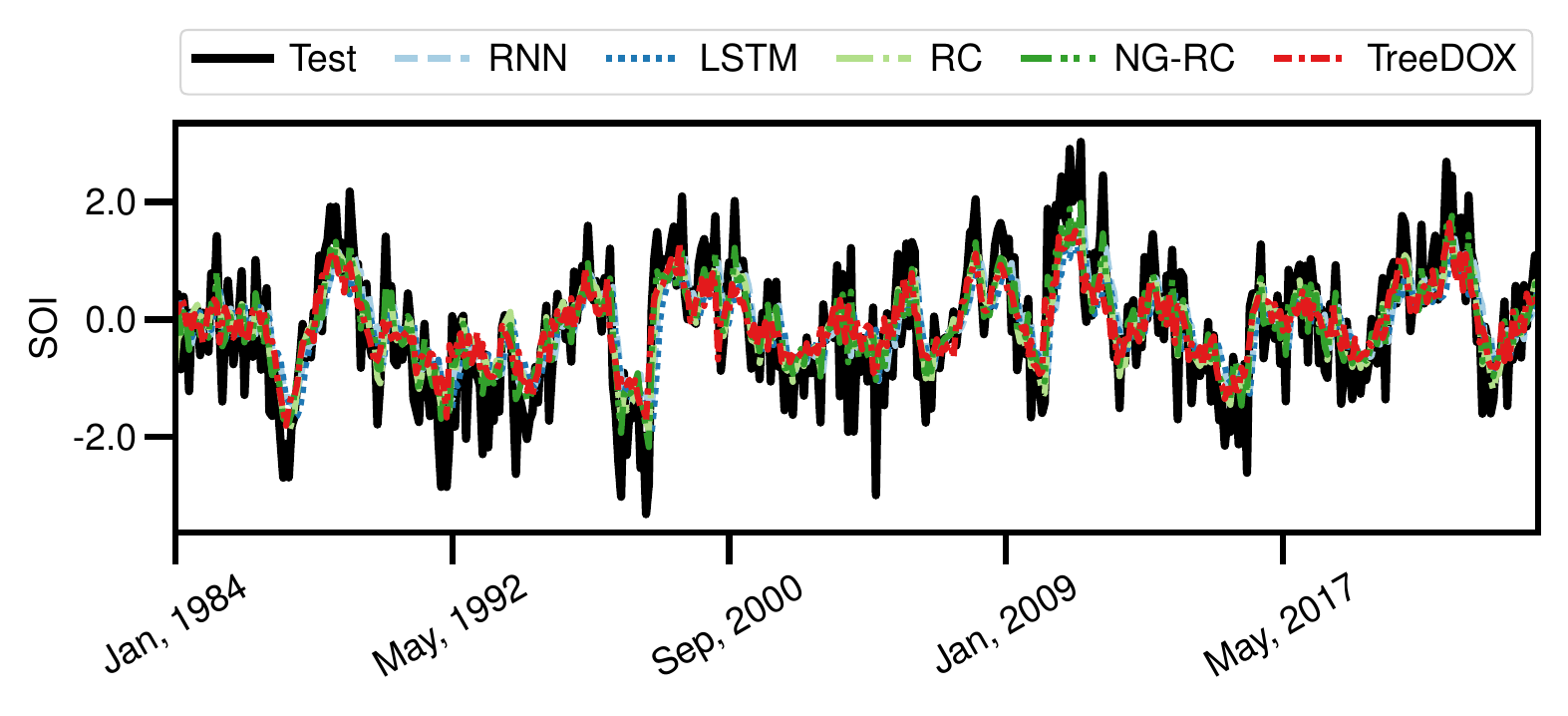}
    \caption{Example 1-month lead SOI forecasts for TreeDOX and other models.}
    \label{fig:soi_forecasts}
\end{figure}

Southern Oscillation Index (SOI) is a useful climate index defined as the z-score of the monthly pressure difference between Tahiti and Darwin \cite{ropelewski1987extension}.
Supplementary Fig.~\ref{fig:soi_raw} displays Southern Oscillation Index data used in the training and testing of TreeDOX and other models to which it is compared in the main manuscript, while Supplementary Fig.~\ref{fig:soi_forecasts} portrays example 1-month lead SOI forecasts associated with Fig. 6 in the main manuscript. 

\begin{figure}
\centering
    \includegraphics[width=\linewidth]{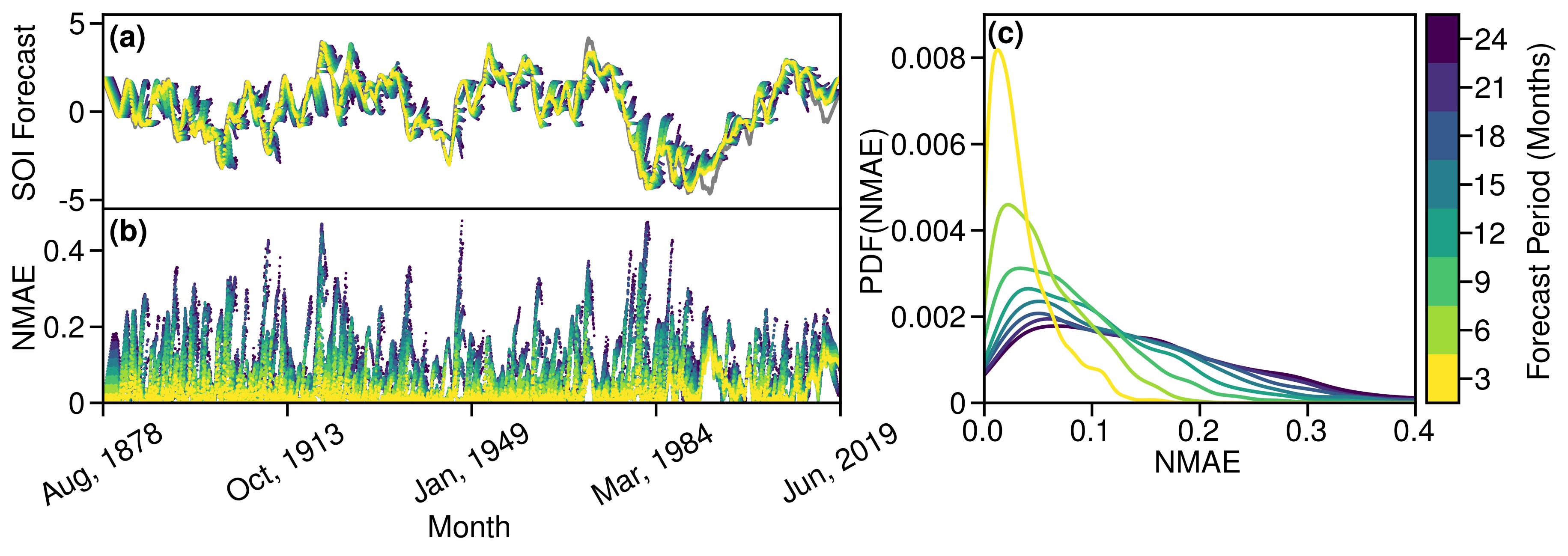}
    \caption{A summary of k-fold cross-validation on SOI time series, where windows of various lengths (3 months, 6 months, etc.) are chosen and data outside the window is used to train TreeDOX, which then performs testing inside the window. \textbf{(a)} SOI tests, where the true data is gray, and the tests are colored according to how long the testing window is. \textbf{(b)} Normalized Mean Absolute Error (NMAE) for each point in each testing window. \textbf{(c)} Summary statistics for the last point in the testing window for each testing window length across all windows using kernel density estimation.}
    \label{fig:soigeneralization}
\end{figure}

We also perform self-evolving prediction on the SOI data to demonstrate a more practical use case. However, due to the noisy nature of SOI data, we apply a Savitzky–Golay filter to reduce noise (and capture the underlying long-term dynamics) in the dataset with a window size of 121 months and a polynomial order of 1 \cite{savitzky_golay_1964}. We also aim to demonstrate the generalizability of TreeDOX and thus perform a rudimentary k-fold cross-validation. In this process, we select a variety of windows in the SOI time series and use all data outside the window to train TreeDOX. We then perform self-evolved predictions for the window. We also vary the length of these windows (from 3 months to 24 months in increments of 3 months) in order to investigate the effect of long predictions for highly chaotic time series data. To summarize the error of these predictions, we use Normalized Mean Absolute Error: $$NMAE(x,\tilde{x}) = \frac{\frac{1}{N}\sum_{i=1}^N |x - \tilde{x}_i|}{\max(x_{train})-\min(x_{train})},$$ where $x$ is the true value, $\tilde{x}$ is a set of forecast realizations, $N$ is the number of forecast realizations, and $x_{train}$ is the training data. Since the SOI data is bounded and ETRs cannot output predictions outside the training range, NMAE captures the fact that the worst possible forecast is where the true and forecasted data are on opposite ends of the bounded range. Therefore, one may interpret an NMAE of 0 to be a perfect forecast and an NMAE of 1 to be the worst possible forecast. Supplementary Fig.~\ref{fig:soigeneralization} displays the results of this experiment, where the NMAE tends to stay below 0.2 for forecast windows less than a full year.  


\section{H\`enon Map}
In addition to the H\`enon map attractor seen in the main manuscript, we wish to investigate the performance of TreeDOX versus current methods with the least amount of bias possible, as well as the effect of training length on forecasts. Therefore, we forecast on 20 sets of data and show the resulting summary statistics in Supplementary Table~\ref{tab:henon}. Each set of data is generated by varying the initial conditions: $(x_0,y_0)=(0.5+0.005m,0.5+0.005m)$, where $0\leq m \leq 19$ is the index of the dataset. Then, 25,000 and 100 training and testing points are created by evolving the H\`enon map. When the training length is varied (as in Supplementary Table~\ref{tab:henon}), the actual training and testing data are fixed, but each model is fed the right-most time window (with the specified training length) of the total 25,000 training points.

Lastly, we perform a sensitivity experiment on H\`enon map data to investigate the efficacy of our hyperparameter prescription procedure. 20 realizations of H\`enon map trajectories are created by evolving the iteration scheme with $(x_0,y_0)=(0.5+0.005m,0.5+0.005m)$, where $0\leq m \leq 19$ is the index of the realization. Then, the three hyperparameters introduced by TreeDOX (namely the delay overembedding dimension, $k$, the delay overembedding lag, $\xi$, and the number of features to use in final training, $p$) are varied similarly to the SOI sensitivity experiment in the main manuscript. More specifically, TreeDOX uses the training data to prescribe hyperparameters as usual, but each hyperparameter is individually varied by applying a multiplicative factor to the prescribed value. See Supplementary Fig.~\ref{fig:henon_sensitivity} for the results of the hyperparameter sensitivity experiment.

\renewcommand{\arraystretch}{1.25}
\begin{table}
    \centering
    \begin{tabular}{C{4em}|c||c|c|c|c|c|}
        \hline
        Training Length & Metric & RNN  & LSTM & RC & NGRC & TreeDOX\\
        \hline\hline
        \multirow{6}{*}{2,500} 
            & RMSE &  $\bm{0.493 \pm 0.056}$  &  $0.523 \pm 0.073$  &  $52.724 \pm 220.856$  &  $0.605 \pm 0.056$  &  $0.606 \pm 0.059$   \\\cline{2-7}
            & NAMI &  $0.064 \pm 0.061$  &  $0.106 \pm 0.080$  &  $0.168 \pm 0.076$  &  $0.242 \pm 0.066$  &  $\bm{0.250 \pm 0.054}$   \\\cline{2-7}
            & Forecast Horizon &  $1.6 \pm 2.7$  &  $1.6 \pm 2.8$  &  $1.1 \pm 2.0$  &  $\bm{21.2 \pm 9.8}$  &  $16.9 \pm 6.7$   \\\cline{2-7}
            & Train+Test Time (s) &  $4.425 \pm 1.115$  &  $5.309 \pm 0.431$  &  $0.463 \pm 0.120$  &  $\bm{0.240 \pm 0.004}$  &  $1.115 \pm 0.075$   \\\cline{2-7}
            & Tune Time (s) &  $45.155 \pm 2.875$  &  $51.693 \pm 2.453$  &  $10.274 \pm 1.086$  &  $\bm{6.103 \pm 0.059}$   & \\\cline{2-7}
            & Total Time (s) &  $49.580 \pm 3.418$  &  $57.002 \pm 2.491$  &  $10.737 \pm 1.107$  &  $6.343 \pm 0.062$  &  $\bm{1.115 \pm 0.075}$   \\\cline{2-7}
        \hline\hline
        \multirow{6}{*}{10,000} 
            & RMSE &  $0.535 \pm 0.075$  &  $\bm{0.516 \pm 0.080}$  &  $201.768 \pm 400.122$  &  $0.567 \pm 0.047$  &  $0.583 \pm 0.050$   \\\cline{2-7}
            & NAMI &  $0.108 \pm 0.082$  &  $0.100 \pm 0.068$  &  $0.145 \pm 0.071$  &  $\bm{0.278 \pm 0.054}$  &  $0.247 \pm 0.029$   \\\cline{2-7}
            & Forecast Horizon &  $1.4 \pm 2.6$  &  $1.6 \pm 2.5$  &  $1.6 \pm 2.8$  &  $\bm{27.7 \pm 9.6}$  &  $19.3 \pm 7.2$   \\\cline{2-7}
            & Train+Test Time (s) &  $5.861 \pm 3.177$  &  $8.864 \pm 5.163$  &  $1.283 \pm 0.365$  &  $\bm{0.481 \pm 0.006}$  &  $1.339 \pm 0.168$   \\\cline{2-7}
            & Tune Time (s) &  $97.073 \pm 2.553$  &  $126.192 \pm 3.609$  &  $27.033 \pm 1.867$  &  $\bm{12.013 \pm 0.115}$   & \\\cline{2-7}
            & Total Time (s) &  $102.935 \pm 4.328$  &  $135.056 \pm 5.311$  &  $28.316 \pm 1.981$  &  $12.494 \pm 0.117$  &  $\bm{1.339 \pm 0.168}$   \\\cline{2-7}
        \hline\hline
        \multirow{6}{*}{17,500} 
            & RMSE &  $0.569 \pm 0.079$  &  $\bm{0.561 \pm 0.097}$  &  $3.818 \pm 6.089$  &  $0.569 \pm 0.069$  &  $0.597 \pm 0.039$   \\\cline{2-7}
            & NAMI &  $0.137 \pm 0.068$  &  $0.133 \pm 0.074$  &  $0.174 \pm 0.072$  &  $\bm{0.284 \pm 0.051}$  &  $0.255 \pm 0.045$   \\\cline{2-7}
            & Forecast Horizon &  $1.4 \pm 3.3$  &  $1.1 \pm 1.6$  &  $1.4 \pm 2.1$  &  $\bm{28.4 \pm 11.2}$  &  $21.6 \pm 5.6$   \\\cline{2-7}
            & Train+Test Time (s) &  $8.491 \pm 4.148$  &  $11.516 \pm 5.186$  &  $2.266 \pm 0.579$  &  $\bm{0.718 \pm 0.011}$  &  $1.598 \pm 0.089$   \\\cline{2-7}
            & Tune Time (s) &  $146.146 \pm 3.273$  &  $199.537 \pm 6.219$  &  $45.477 \pm 2.605$  &  $\bm{19.648 \pm 1.787}$   & \\\cline{2-7}
            & Total Time (s) &  $154.637 \pm 5.684$  &  $211.053 \pm 7.965$  &  $47.743 \pm 2.681$  &  $20.365 \pm 1.785$  &  $\bm{1.598 \pm 0.089}$   \\\cline{2-7}
        \hline\hline
        \multirow{6}{*}{25,000}
            & RMSE &  $0.562 \pm 0.095$  &  $\bm{0.545 \pm 0.089}$  &  $107.804 \pm 312.731$  &  $0.546 \pm 0.047$  &  $0.584 \pm 0.058$   \\\cline{2-7}
            & NAMI &  $0.118 \pm 0.077$  &  $0.130 \pm 0.076$  &  $0.158 \pm 0.078$  &  $\bm{0.297 \pm 0.050}$  &  $0.247 \pm 0.044$   \\\cline{2-7}
            & Forecast Horizon &  $1.3 \pm 2.0$  &  $1.4 \pm 2.0$  &  $1.2 \pm 2.4$  &  $\bm{31.0 \pm 9.2}$  &  $22.6 \pm 7.7$   \\\cline{2-7}
            & Train+Test Time (s) &  $10.230 \pm 6.182$  &  $16.396 \pm 10.151$  &  $3.189 \pm 0.743$  &  $\bm{0.960 \pm 0.016}$  &  $1.827 \pm 0.088$   \\\cline{2-7}
            & Tune Time (s) &  $196.961 \pm 5.309$  &  $271.967 \pm 6.392$  &  $62.555 \pm 3.233$  &  $\bm{23.712 \pm 0.113}$   & \\\cline{2-7}
            & Total Time (s) &  $207.192 \pm 8.990$  &  $288.363 \pm 11.327$  &  $65.744 \pm 3.325$  &  $24.672 \pm 0.117$  &  $\bm{1.827 \pm 0.088}$   \\\cline{2-7}
        \hline
    \end{tabular}
    \caption{Numerical experiment with varying training length and 20 total H\`enon map independent trajectories. Each cell displays the mean metric across the 20 realizations, plus or minus their standard deviations. The bold cells represent the best model in the respective category. RMSE is Root Mean Square Error, NAMI is Normalized Average Mutual Information, and Forecast Horizon represents the number of time points for which the absolute residual error for all dimensions is less than the standard deviation of the training data for that respective coordinate. The forecast horizon represents the length of time for which the forecasts are qualitatively converged to the test data. Hyperparameter tuning for RNN, LSTM, RC, and NGRC uses TPE optimization (see the main manuscript) with 25 samples.}
    \label{tab:henon}
\end{table}

\begin{figure}
    \centering
    \includegraphics[width=\linewidth]{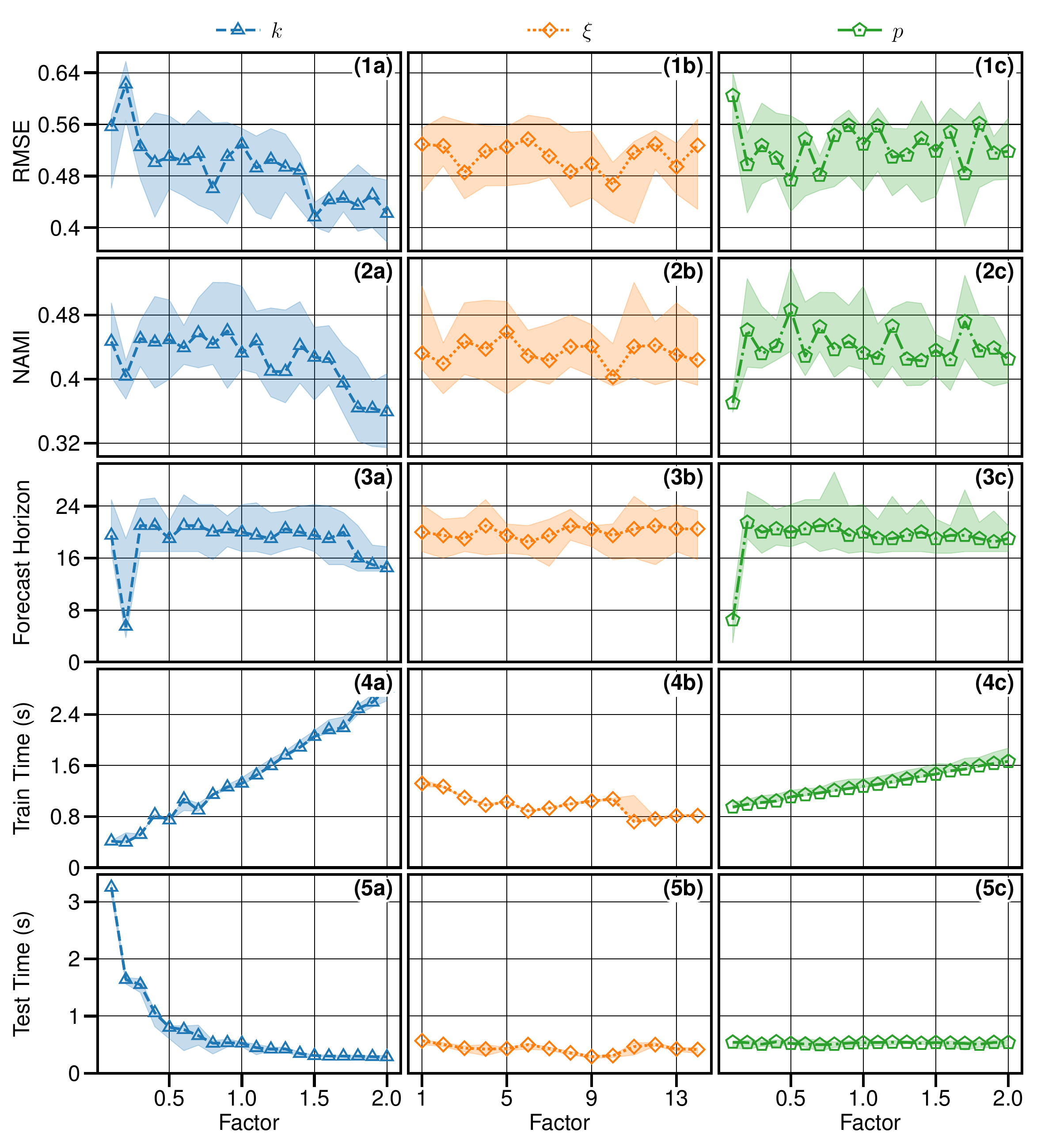}
    \caption{Sensitivity experiment results for the H\`enon map, where 20 realizations of H\`enon map trajectories are generated, and the three hyperparameters introduced by TreeDOX (namely the delay overembedding dimension, $k$, the delay overembedding lag, $\xi$, and the number of features to use in final training, $p$) are varied via a multiplicative factor applied to the value prescribed through the usual technique using training data. Therefore, a factor of 1.0 implies the respective hyperparameter is equal to the default prescribed value.}
    \label{fig:henon_sensitivity}
\end{figure}

\section{Lorenz System}
See Supplementary Fig.~\ref{fig:D2} for visualizations of Grassberger-Procaccia algorithm \cite{grassberger1983characterization} results to estimate the correlation dimension, $D_2$, of the test and predicted Lorenz attractors, respectively. To further demonstrate the ability of TreeDOX to forecast the `climate' (as in the long-term dynamics) in addition to the Lorenz attractor recreation in the main manuscript, we verify the long-term forecasts with a recreation of the return map of the $z$ coordinate. More specifically, there exists a relationship between successive peaks in the $z$ coordinate of Lorenz trajectories. Here, $M_i$ corresponds to the $i$-th local maxima of the $z$ coordinate signal, and the return map is generated by plotting $M_{i+1}$ against $M_i$. See Supplementary Fig.~\ref{fig:return_map_vis} and Supplementary Fig.~\ref{fig:return_map} for results.

\begin{figure}
    \centering
    \includegraphics[width=\linewidth]{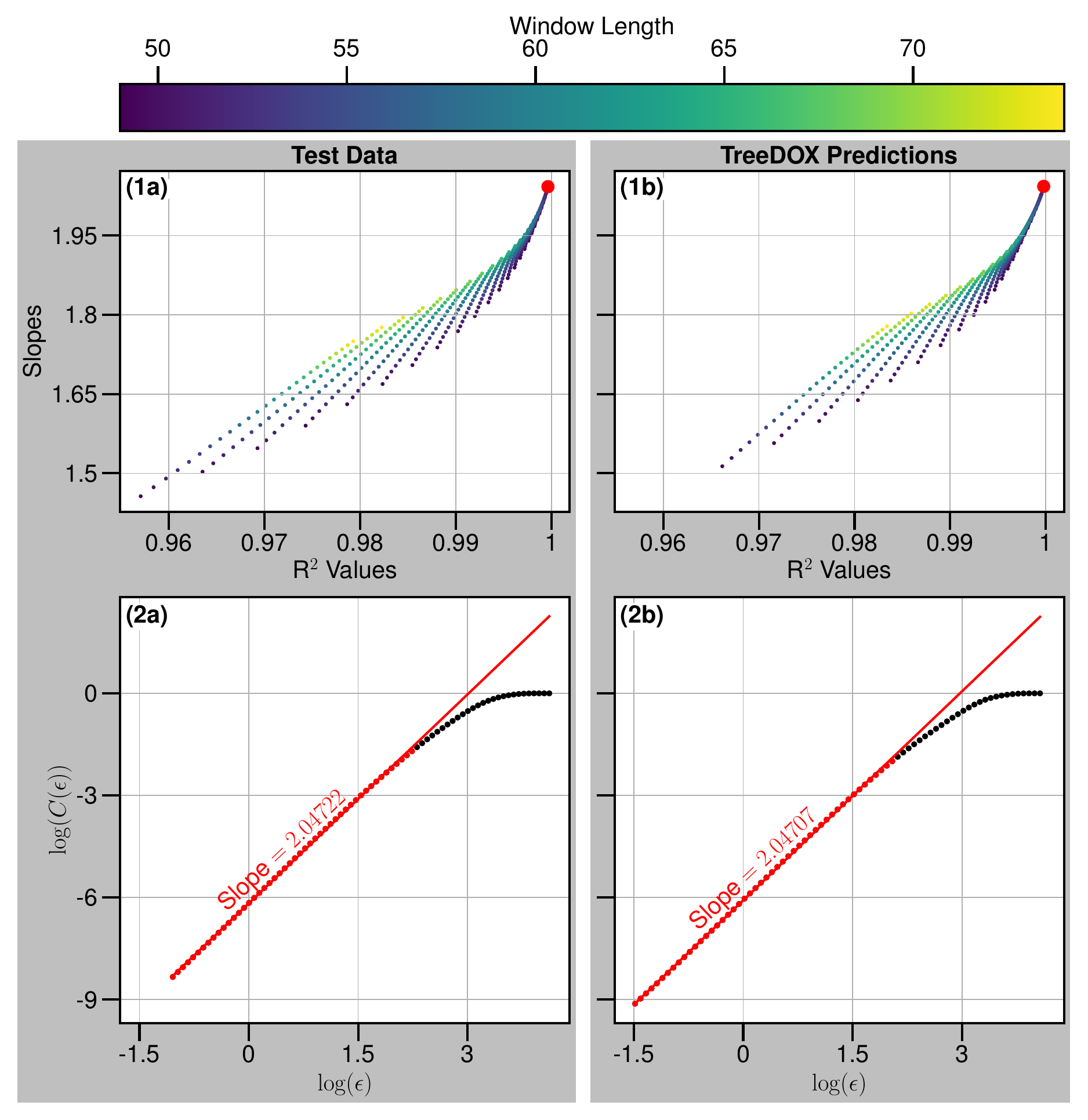}
    \caption{Correlation dimension, $D_2$, of test and predicted Lorenz attractor data. Here, we use the Grassberger-Procaccia algorithm \cite{grassberger1983characterization} and find the contiguous window of $\varepsilon$ whose coefficient of restitution of linear regression is highest, the resulting slope being taken as the corresponding $D_2$.}
    \label{fig:D2}
\end{figure}

\begin{figure}
    \centering
    \includegraphics[width=\linewidth]{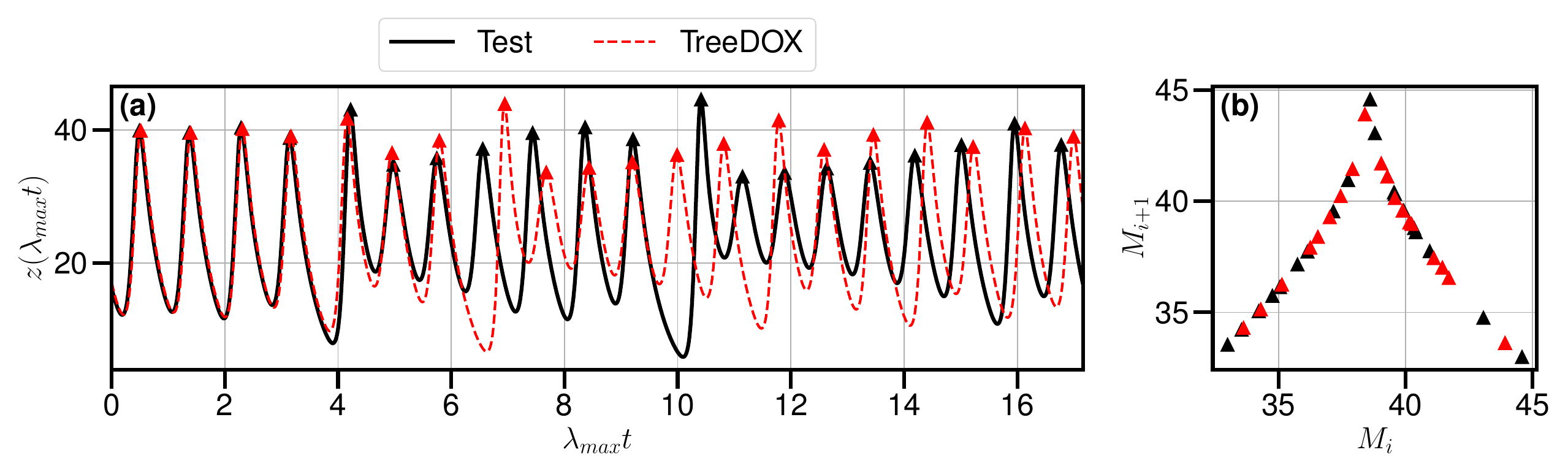}
    \caption{A visualization of return map data, where the test and predicted data shown in \textbf{(a)} are the first 1,500 points in the Lorenz attractor plot seen in the main manuscript. \textbf{(b)} displays the return map resulting from plotting each peak in the signal against the previous peak, where black and red are the test and TreeDOX forecast, respectively.}
    \label{fig:return_map_vis}
\end{figure}

\sidecaptionvpos{figure}{c}
\begin{SCfigure}[50]
    \centering
    \includegraphics[width=0.5\linewidth]{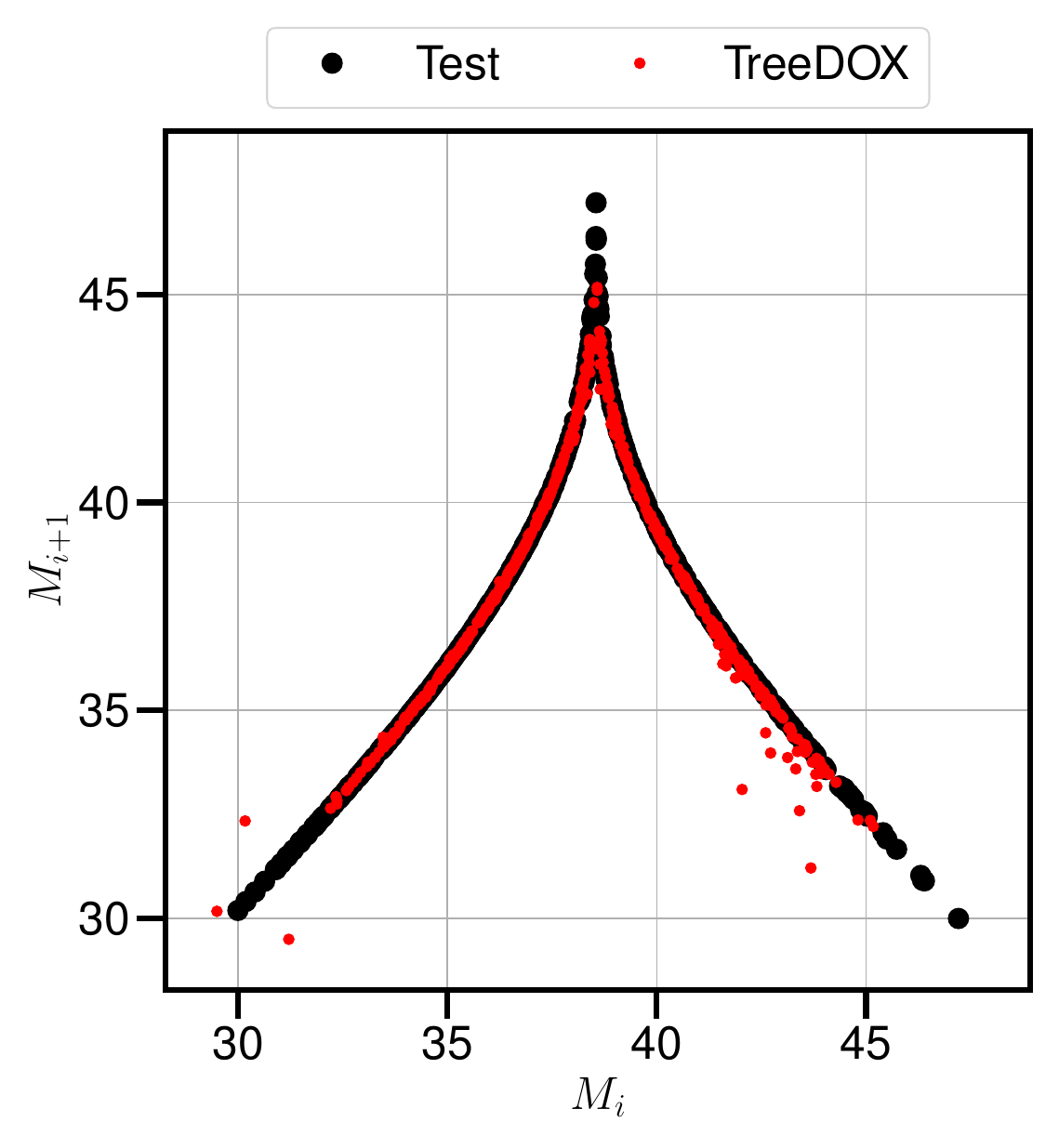}
    \caption{Final return map from the full trajectories seen in the Lorenz attractor figure seen in the main manuscript. Here, black and red dots represent the test data and TreeDOX forecasts, respectively.}
    \label{fig:return_map}
\end{SCfigure}

Similar to the previous section, we wish to investigate the performance of TreeDOX versus current methods on Lorenz system forecasts with the least amount of bias possible, as well as the effect of training length on forecasts. Therefore, we forecast on 20 sets of data and show the resulting summary statistics in Supplementary Table~\ref{tab:lorenz}. Each set of data is generated by varying the initial conditions: $(x(t=0),y(t=0),z(t=0))=(1+0.01m,1+0.01m,1+0.01m)$, where $0\leq m \leq 19$ is the index of the dataset. Then, 25,000 and 1,500 training and testing points are created by evolving the Lorenz system with RK45 and $dt=0.01$. When the training length is varied (as in Supplementary Table~\ref{tab:lorenz}), the actual training and testing data are fixed, but each model is fed the right-most time window (with the specified training length) of the total 25,000 training points.

Furthermore, we investigate the effect of noise on forecasts. We repeat the previous experiment but keep the training length fixed at 25,000 and instead vary the scale of additional white noise. We introduce a noise factor, $\nu$, where white noise is drawn from $N(0,\nu\sigma_i)$ (normal distribution with mean of zero and standard deviation of $\nu\sigma_i$, where $\sigma_i$ is the standard deviation of the $i$-th coordinate of the training data) and added to the $i$-th coordinate of training data. The testing data is left noiseless to allow for direct comparison of forecast performance between noise levels. See Supplementary Table~\ref{tab:lorenz_noise} for the results of the numerical experiment.

Lastly, we perform a sensitivity experiment on H\`enon map data to investigate the efficacy of our hyperparameter prescription procedure. 20 realizations of H\`enon map trajectories are created by evolving the iteration scheme with $(x_0,y_0)=(0.5+0.005m,0.5+0.005m)$, where $0\leq m \leq 19$ is the index of the realization. Then, the three hyperparameters introduced by TreeDOX (namely the delay overembedding dimension, $k$, the delay overembedding lag, $\xi$, and the number of features to use in final training, $p$) are varied similarly to the SOI sensitivity experiment in the main manuscript. More specifically, TreeDOX uses the training data to prescribe hyperparameters as usual, but each hyperparameter is individually varied by applying a multiplicative factor to the prescribed value. See Supplementary Fig.~\ref{fig:lorenz_sensitivity} for the results of the hyperparameter sensitivity experiment.

\renewcommand{\arraystretch}{1.25}
\begin{table}
    \centering
    \begin{tabular}{C{4em}|c||c|c|c|c|c|}
        \hline
        Training Length & Metric & RNN  & LSTM & RC & NGRC & TreeDOX\\
        \hline\hline
        \multirow{6}{*}{2,500} 
            & RMSE &  $9.056 \pm 0.750$  &  $\bm{8.835 \pm 0.452}$  &  $10.042 \pm 1.331$  &  $10.111 \pm 1.372$  &  $10.141 \pm 1.048$   \\\cline{2-7}
            & NAMI &  $0.010 \pm 0.009$  &  $0.007 \pm 0.010$  &  $0.159 \pm 0.078$  &  $\bm{0.176 \pm 0.098}$  &  $0.129 \pm 0.053$   \\\cline{2-7}
            & Forecast Horizon &  $5.8 \pm 9.5$  &  $6.5 \pm 11.3$  &  $324.6 \pm 220.3$  &  $\bm{357.1 \pm 231.9}$  &  $245.6 \pm 96.2$   \\\cline{2-7}
            & Train+Test Time (s) &  $19.047 \pm 3.988$  &  $26.295 \pm 7.713$  &  $0.633 \pm 0.302$  &  $\bm{0.291 \pm 0.013}$  &  $2.512 \pm 0.542$   \\\cline{2-7}
            & Tune Time (s) &  $242.704 \pm 54.409$  &  $344.463 \pm 91.989$  &  $11.579 \pm 1.943$  &  $\bm{6.938 \pm 1.734}$   & \\\cline{2-7}
            & Total Time (s) &  $261.751 \pm 58.394$  &  $370.758 \pm 98.365$  &  $12.212 \pm 1.989$  &  $7.229 \pm 1.737$  &  $\bm{2.512 \pm 0.542}$   \\\cline{2-7}
        \hline\hline
        \multirow{6}{*}{10,000} 
            & RMSE &  $11.381 \pm 1.247$  &  $11.330 \pm 1.062$  &  $10.193 \pm 1.104$  &  $\bm{9.672 \pm 1.317}$  &  $10.007 \pm 0.920$   \\\cline{2-7}
            & NAMI &  $0.053 \pm 0.029$  &  $0.044 \pm 0.020$  &  $0.164 \pm 0.069$  &  $\bm{0.171 \pm 0.077}$  &  $0.168 \pm 0.049$   \\\cline{2-7}
            & Forecast Horizon &  $68.5 \pm 70.0$  &  $50.9 \pm 24.9$  &  $330.9 \pm 190.9$  &  $\bm{397.6 \pm 217.5}$  &  $389.9 \pm 124.0$   \\\cline{2-7}
            & Train+Test Time (s) &  $19.412 \pm 0.637$  &  $33.312 \pm 11.599$  &  $1.632 \pm 0.771$  &  $\bm{0.670 \pm 0.128}$  &  $6.214 \pm 0.818$   \\\cline{2-7}
            & Tune Time (s) &  $269.381 \pm 55.484$  &  $390.069 \pm 94.008$  &  $30.606 \pm 3.306$  &  $\bm{16.057 \pm 1.674}$   & \\\cline{2-7}
            & Total Time (s) &  $288.792 \pm 55.760$  &  $423.381 \pm 103.806$  &  $32.238 \pm 3.468$  &  $16.726 \pm 1.649$  &  $\bm{6.214 \pm 0.818}$   \\\cline{2-7}
        \hline\hline
        \multirow{6}{*}{17,500} 
            & RMSE &  $11.917 \pm 1.127$  &  $12.018 \pm 0.716$  &  $10.351 \pm 1.032$  &  $9.802 \pm 1.496$  &  $\bm{9.435 \pm 1.279}$   \\\cline{2-7}
            & NAMI &  $0.058 \pm 0.031$  &  $0.054 \pm 0.019$  &  $0.142 \pm 0.062$  &  $0.176 \pm 0.087$  &  $\bm{0.187 \pm 0.075}$   \\\cline{2-7}
            & Forecast Horizon &  $86.2 \pm 73.7$  &  $95.0 \pm 71.0$  &  $272.6 \pm 174.4$  &  $311.8 \pm 217.0$  &  $\bm{464.5 \pm 185.5}$   \\\cline{2-7}
            & Train+Test Time (s) &  $20.312 \pm 0.765$  &  $38.265 \pm 14.439$  &  $2.673 \pm 0.924$  &  $\bm{1.128 \pm 0.282}$  &  $10.583 \pm 0.780$   \\\cline{2-7}
            & Tune Time (s) &  $280.912 \pm 11.249$  &  $458.587 \pm 110.341$  &  $50.710 \pm 5.192$  &  $\bm{25.117 \pm 2.056}$   & \\\cline{2-7}
            & Total Time (s) &  $301.223 \pm 11.957$  &  $496.852 \pm 119.718$  &  $53.382 \pm 5.444$  &  $26.245 \pm 2.044$  &  $\bm{10.583 \pm 0.780}$   \\\cline{2-7}
        \hline\hline
        \multirow{6}{*}{25,000}
            & RMSE &  $11.898 \pm 1.180$  &  $11.871 \pm 0.770$  &  $9.816 \pm 1.393$  &  $9.886 \pm 1.406$  &  $\bm{9.487 \pm 1.222}$   \\\cline{2-7}
            & NAMI &  $0.056 \pm 0.022$  &  $0.068 \pm 0.048$  &  $0.160 \pm 0.070$  &  $0.169 \pm 0.100$  &  $\bm{0.212 \pm 0.086}$   \\\cline{2-7}
            & Forecast Horizon &  $113.2 \pm 100.1$  &  $92.8 \pm 61.0$  &  $331.4 \pm 190.7$  &  $328.4 \pm 250.9$  &  $\bm{455.9 \pm 149.9}$   \\\cline{2-7}
            & Train+Test Time (s) &  $22.988 \pm 4.999$  &  $41.006 \pm 15.529$  &  $3.925 \pm 1.440$  &  $\bm{1.409 \pm 0.327}$  &  $15.434 \pm 1.715$   \\\cline{2-7}
            & Tune Time (s) &  $311.664 \pm 31.783$  &  $531.519 \pm 113.487$  &  $68.359 \pm 5.857$  &  $\bm{34.001 \pm 2.961}$   & \\\cline{2-7}
            & Total Time (s) &  $334.653 \pm 36.211$  &  $572.525 \pm 121.025$  &  $72.284 \pm 6.051$  &  $35.410 \pm 2.933$  &  $\bm{15.434 \pm 1.715}$   \\\cline{2-7}
        \hline
    \end{tabular}
    \caption{Numerical experiment with varying training length and 20 total Lorenz system independent trajectories. Each cell displays the mean metric across the 20 realizations, plus or minus their standard deviations. The bold cells represent the best model in the respective category. RMSE is Root Mean Square Error, NAMI is Normalized Average Mutual Information, and Forecast Horizon represents the number of time points for which the absolute residual error for all dimensions is less than the standard deviation of the training data for that respective coordinate. The forecast horizon represents the length of time for which the forecasts are qualitatively converged to the test data. Hyperparameter tuning for RNN, LSTM, RC, and NGRC uses TPE optimization (see the main manuscript) with 25 samples.}
    \label{tab:lorenz}
\end{table}

\renewcommand{\arraystretch}{1.25}
\begin{table}
    \centering
    \begin{tabular}{C{4em}|c||c|c|c|c|c|}
        \hline
        Noise Factor, $\nu$ & Metric & RNN  & LSTM & RC & NGRC & TreeDOX\\
        \hline\hline
        \multirow{6}{*}{0.001} 
            & RMSE &  $11.904 \pm 1.179$  &  $11.838 \pm 0.757$  &  $\bm{9.017 \pm 1.862}$  &  $9.315 \pm 1.623$  &  $9.442 \pm 1.222$   \\\cline{2-7}
            & NAMI &  $0.057 \pm 0.022$  &  $0.067 \pm 0.049$  &  $\bm{0.221 \pm 0.126}$  &  $0.217 \pm 0.123$  &  $0.210 \pm 0.112$   \\\cline{2-7}
            & Forecast Horizon &  $113.3 \pm 100.0$  &  $92.8 \pm 61.0$  &  $\bm{490.9 \pm 253.8}$  &  $421.7 \pm 180.6$  &  $484.6 \pm 191.8$   \\\cline{2-7}
            & Train+Test Time (s) &  $23.338 \pm 5.494$  &  $30.733 \pm 10.489$  &  $3.634 \pm 1.806$  &  $\bm{1.285 \pm 0.278}$  &  $14.968 \pm 1.361$   \\\cline{2-7}
            & Tune Time (s) &  $331.279 \pm 84.302$  &  $388.747 \pm 20.842$  &  $68.327 \pm 4.531$  &  $\bm{33.950 \pm 2.231}$   & \\\cline{2-7}
            & Total Time (s) &  $354.617 \pm 89.656$  &  $419.480 \pm 21.120$  &  $71.961 \pm 5.228$  &  $35.235 \pm 2.230$  &  $\bm{14.968 \pm 1.361}$   \\\cline{2-7}
        \hline\hline
        \multirow{6}{*}{0.01} 
            & RMSE &  $12.012 \pm 1.187$  &  $11.758 \pm 0.817$  &  $\bm{8.996 \pm 1.385}$  &  $9.614 \pm 1.085$  &  $9.646 \pm 1.051$   \\\cline{2-7}
            & NAMI &  $0.058 \pm 0.026$  &  $0.076 \pm 0.053$  &  $\bm{0.226 \pm 0.075}$  &  $0.190 \pm 0.073$  &  $0.217 \pm 0.108$   \\\cline{2-7}
            & Forecast Horizon &  $112.1 \pm 98.6$  &  $92.1 \pm 60.8$  &  $434.1 \pm 167.3$  &  $384.1 \pm 160.9$  &  $\bm{447.0 \pm 182.7}$   \\\cline{2-7}
            & Train+Test Time (s) &  $21.540 \pm 1.267$  &  $30.709 \pm 10.527$  &  $3.059 \pm 1.342$  &  $\bm{1.351 \pm 0.318}$  &  $15.008 \pm 1.974$   \\\cline{2-7}
            & Tune Time (s) &  $313.366 \pm 30.081$  &  $388.138 \pm 21.179$  &  $67.454 \pm 4.518$  &  $\bm{33.668 \pm 1.529}$   & \\\cline{2-7}
            & Total Time (s) &  $334.906 \pm 30.036$  &  $418.847 \pm 21.209$  &  $70.513 \pm 4.995$  &  $35.019 \pm 1.601$  &  $\bm{15.008 \pm 1.974}$   \\\cline{2-7}
        \hline\hline
        \multirow{6}{*}{0.1} 
            & RMSE &  $11.923 \pm 0.906$  &  $11.688 \pm 0.855$  &  $10.444 \pm 0.745$  &  $10.826 \pm 0.787$  &  $\bm{9.968 \pm 1.115}$   \\\cline{2-7}
            & NAMI &  $0.053 \pm 0.018$  &  $0.067 \pm 0.041$  &  $\bm{0.142 \pm 0.057}$  &  $0.090 \pm 0.044$  &  $0.128 \pm 0.063$   \\\cline{2-7}
            & Forecast Horizon &  $75.4 \pm 40.4$  &  $73.4 \pm 52.0$  &  $257.5 \pm 125.6$  &  $174.7 \pm 111.1$  &  $\bm{287.3 \pm 154.4}$   \\\cline{2-7}
            & Train+Test Time (s) &  $23.164 \pm 5.981$  &  $30.259 \pm 10.936$  &  $3.019 \pm 1.006$  &  $\bm{1.478 \pm 0.287}$  &  $16.188 \pm 2.416$   \\\cline{2-7}
            & Tune Time (s) &  $311.331 \pm 29.620$  &  $387.955 \pm 21.589$  &  $65.497 \pm 4.388$  &  $\bm{33.912 \pm 1.856}$   & \\\cline{2-7}
            & Total Time (s) &  $334.495 \pm 33.417$  &  $418.214 \pm 23.926$  &  $68.516 \pm 4.601$  &  $35.390 \pm 1.916$  &  $\bm{16.188 \pm 2.416}$   \\\cline{2-7}
        \hline\hline
        \multirow{6}{*}{1.0}
            & RMSE &  $8.751 \pm 0.775$  &  $\bm{8.695 \pm 0.664}$  &  $34.788 \pm 26.359$  &  $10.094 \pm 1.067$  &  $9.855 \pm 0.919$   \\\cline{2-7}
            & NAMI &  $0.004 \pm 0.002$  &  $0.008 \pm 0.017$  &  $\bm{0.094 \pm 0.027}$  &  $0.053 \pm 0.015$  &  $0.062 \pm 0.023$   \\\cline{2-7}
            & Forecast Horizon &  $7.2 \pm 10.9$  &  $6.5 \pm 10.5$  &  $2.7 \pm 10.5$  &  $32.2 \pm 23.7$  &  $\bm{47.8 \pm 32.8}$   \\\cline{2-7}
            & Train+Test Time (s) &  $22.394 \pm 5.498$  &  $21.293 \pm 0.227$  &  $1.978 \pm 0.670$  &  $\bm{1.501 \pm 0.223}$  &  $14.469 \pm 3.653$   \\\cline{2-7}
            & Tune Time (s) &  $325.516 \pm 69.582$  &  $388.331 \pm 20.158$  &  $62.799 \pm 3.334$  &  $\bm{33.707 \pm 1.656}$   & \\\cline{2-7}
            & Total Time (s) &  $347.910 \pm 75.062$  &  $409.624 \pm 20.261$  &  $64.777 \pm 3.407$  &  $35.209 \pm 1.645$  &  $\bm{14.469 \pm 3.653}$   \\\cline{2-7}
        \hline
    \end{tabular}
    \caption{Numerical experiment with varying noise level and 20 total Lorenz system independent trajectories. Each cell displays the mean metric across the 20 realizations, plus or minus their standard deviations. The bold cells represent the best model in the respective category. RMSE is Root Mean Square Error, NAMI is Normalized Average Mutual Information, and Forecast Horizon represents the number of time points for which the absolute residual error for all dimensions is less than the standard deviation of the training data for that respective coordinate. The forecast horizon represents the length of time for which the forecasts are qualitatively converged to the test data. Hyperparameter tuning for RNN, LSTM, RC, and NGRC uses TPE optimization (see the main manuscript) with 25 samples.}
    \label{tab:lorenz_noise}
\end{table}

\begin{figure}
    \centering
    \includegraphics[width=\linewidth]{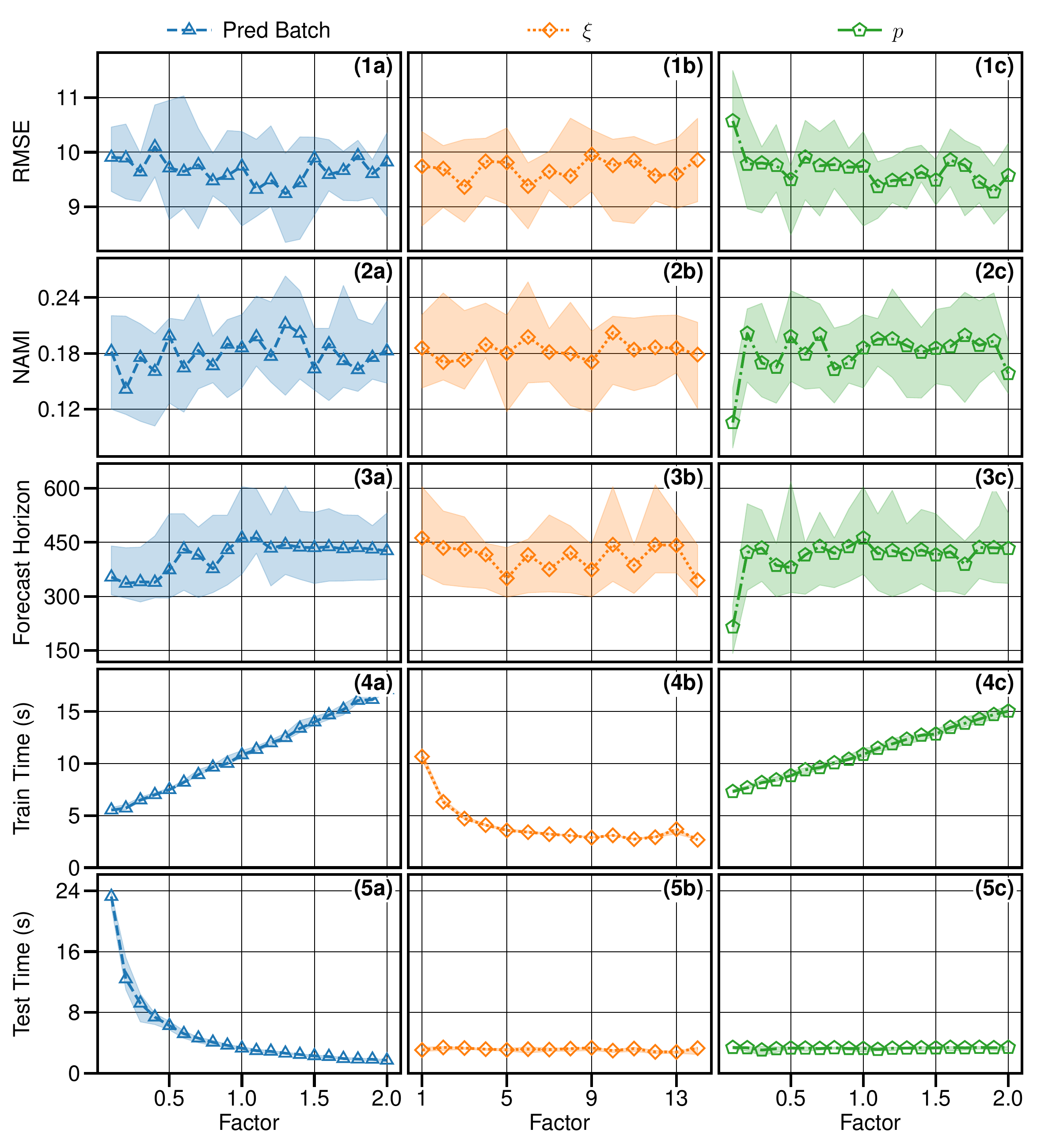}
    \caption{Sensitivity experiment results for the Lorenz system, where 20 realizations of Lorenz system trajectories are generated, and the three hyperparameters introduced by TreeDOX (namely the delay overembedding dimension, $k$, the delay overembedding lag, $\xi$, and the number of features to use in final training, $p$) are varied via a multiplicative factor applied to the value prescribed through the usual technique using training data. Therefore, a factor of 1.0 implies the respective hyperparameter is equal to the default prescribed value.}
    \label{fig:lorenz_sensitivity}
\end{figure}

\section{Double-scroll System}
To further investigate the efficacy of TreeDOX for continuous chaotic dynamics, we perform forecasting on the double-scroll electronic circuit \cite{chang1998stabilizing}:
\begin{align*}
    \dot{V_1} &= \frac{V_1}{R_1} - \frac{V_1-V_2}{R_2} - 2I_r\sinh(\beta(V_1-V_2)) \\
    \dot{V_2} &= \frac{V_1-V_2}{R_2} + 2I_r\sinh(\beta(V_1-V_2)) - I \\
    \dot{I} &= V_2 - R_4I
\end{align*}
where $R_1=1.2$, $R_2=3.44$, $R_4=0.193$, $\beta=11.6$, and $I_r=2.25\times10^{-5}$. Here, we evolve the system with RK45 with $dt=0.25$ and $(V_1,V_2,I)(t=0)=(1,1,1)$, and use 25,000 and 500 training and testing points, respectively. See Supplementary Fig.~\ref{fig:doublescroll} for forecast results.

Additionally, to remove bias from these results and to demonstrate the effect of training length on TreeDOX and other models, we forecast on 20 sets of data and show the resulting summary statistics in Supplementary Table~\ref{tab:doublescroll_results}. Each set of data is generated by varying the initial conditions: $(V_1(t=0),V_2(t=0),I(t=0))=(1+0.01m,1+0.01m,1+0.01m)$, where $0\leq m \leq 19$ is the index of the dataset. Then, 25,000 and 500 training and testing points are created by evolving RK45 with $dt=0.25$. When the training length is varied (as in Supplementary Table~\ref{tab:doublescroll_results}), the actual training and testing data are fixed, but each model is fed the right-most time window (with the specified training length) of the total 25,000 training points.

\begin{figure}
    \centering
    \includegraphics[width=\linewidth]{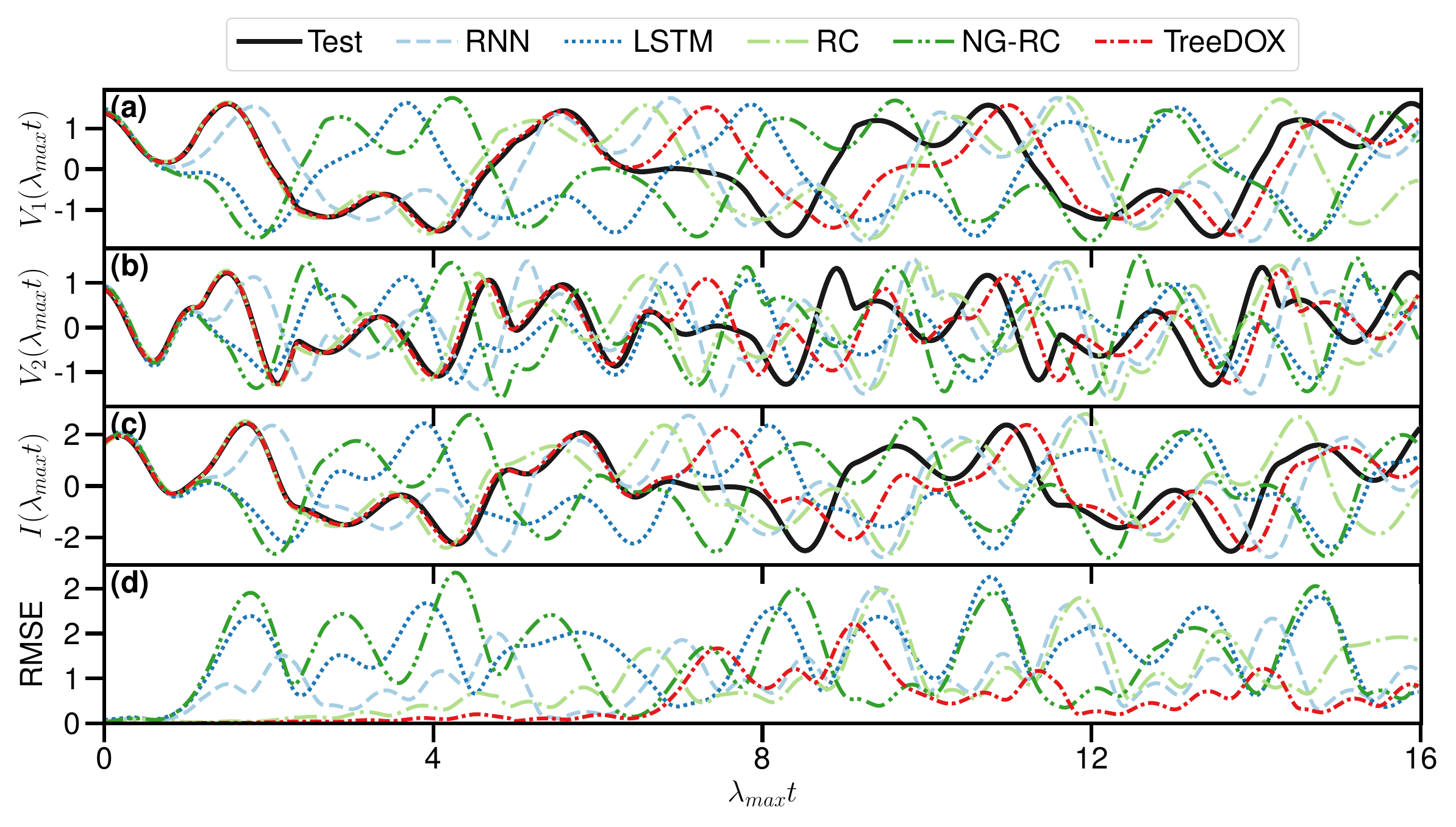}
    \caption{Summary of double-scroll forecasts with $R_1=1.2$, $R_2=3.44$, $R_4=0.193$, $\beta=11.6$, and $I_r=2.25\times10^{-5}$, displayed using Lyapunov time where $\lambda_{max}=7.81$. Test data is black while RNN, LSTM, RC, NG-RC, and TreeDOX are light blue, blue, light green, green, and red, respectively. \textbf{(a,b,c)} $V_1$, $V_2$, and $I$ coordinates respectively. \textbf{(d)} Root Mean Square Error (RMSE) of $V_1$, $V_2$, and $I$ forecasted versus test data. Here, 25,000 and 500 training and testing points were used, respectively.}
    \label{fig:doublescroll}
\end{figure}

\renewcommand{\arraystretch}{1.25}
\begin{table}
    \centering
    \begin{tabular}{C{4em}|c||c|c|c|c|c|}
        \hline
        Training Length & Metric & RNN  & LSTM & RC & NGRC & TreeDOX\\
        \hline\hline
        \multirow{6}{*}{2,500} 
            & RMSE &  $0.870 \pm 0.085$  &  $\bm{0.851 \pm 0.082}$  &  $0.960 \pm 0.173$  &  $0.993 \pm 0.137$  &  $0.890 \pm 0.150$   \\\cline{2-7}
            & NAMI &  $0.065 \pm 0.057$  &  $0.038 \pm 0.040$  &  $0.160 \pm 0.049$  &  $0.154 \pm 0.040$  &  $\bm{0.182 \pm 0.074}$   \\\cline{2-7}
            & Forecast Horizon &  $8.6 \pm 17.2$  &  $6.5 \pm 12.5$  &  $107.4 \pm 55.9$  &  $97.8 \pm 53.0$  &  $\bm{128.4 \pm 73.4}$   \\\cline{2-7}
            & Train+Test Time (s) &  $16.421 \pm 2.789$  &  $9.227 \pm 3.273$  &  $0.388 \pm 0.014$  &  $\bm{0.221 \pm 0.017}$  &  $1.644 \pm 0.133$   \\\cline{2-7}
            & Tune Time (s) &  $202.918 \pm 33.515$  &  $117.475 \pm 35.617$  &  $10.959 \pm 0.411$  &  $\bm{4.992 \pm 0.838}$   & \\\cline{2-7}
            & Total Time (s) &  $219.339 \pm 34.120$  &  $126.702 \pm 38.541$  &  $11.347 \pm 0.417$  &  $5.213 \pm 0.840$  &  $\bm{1.644 \pm 0.133}$   \\\cline{2-7}
        \hline\hline
        \multirow{6}{*}{10,000} 
            & RMSE &  $1.035 \pm 0.126$  &  $1.039 \pm 0.125$  &  $1.194 \pm 0.707$  &  $1.009 \pm 0.140$  &  $\bm{0.811 \pm 0.132}$   \\\cline{2-7}
            & NAMI &  $0.135 \pm 0.032$  &  $0.128 \pm 0.025$  &  $0.166 \pm 0.052$  &  $0.153 \pm 0.053$  &  $\bm{0.217 \pm 0.070}$   \\\cline{2-7}
            & Forecast Horizon &  $56.5 \pm 39.3$  &  $55.2 \pm 32.7$  &  $67.7 \pm 47.4$  &  $109.3 \pm 72.7$  &  $\bm{177.9 \pm 58.1}$   \\\cline{2-7}
            & Train+Test Time (s) &  $19.651 \pm 1.278$  &  $14.226 \pm 5.975$  &  $0.996 \pm 0.119$  &  $\bm{0.542 \pm 0.118}$  &  $6.272 \pm 0.536$   \\\cline{2-7}
            & Tune Time (s) &  $247.023 \pm 14.885$  &  $176.130 \pm 45.488$  &  $27.330 \pm 1.489$  &  $\bm{11.071 \pm 0.283}$   & \\\cline{2-7}
            & Total Time (s) &  $266.674 \pm 15.447$  &  $190.355 \pm 49.884$  &  $28.326 \pm 1.555$  &  $11.614 \pm 0.344$  &  $\bm{6.272 \pm 0.536}$   \\\cline{2-7}
        \hline\hline
        \multirow{6}{*}{17,500} 
            & RMSE &  $1.048 \pm 0.155$  &  $1.049 \pm 0.133$  &  $6.023 \pm 11.942$  &  $0.975 \pm 0.165$  &  $\bm{0.769 \pm 0.179}$   \\\cline{2-7}
            & NAMI &  $0.131 \pm 0.040$  &  $0.149 \pm 0.027$  &  $0.145 \pm 0.039$  &  $0.114 \pm 0.085$  &  $\bm{0.277 \pm 0.108}$   \\\cline{2-7}
            & Forecast Horizon &  $56.9 \pm 46.2$  &  $71.5 \pm 27.9$  &  $73.3 \pm 54.7$  &  $64.7 \pm 47.7$  &  $\bm{220.4 \pm 67.7}$   \\\cline{2-7}
            & Train+Test Time (s) &  $20.228 \pm 3.024$  &  $20.744 \pm 10.440$  &  $1.728 \pm 0.777$  &  $\bm{0.888 \pm 0.205}$  &  $11.454 \pm 0.338$   \\\cline{2-7}
            & Tune Time (s) &  $282.700 \pm 16.729$  &  $243.257 \pm 54.972$  &  $42.277 \pm 1.900$  &  $\bm{17.436 \pm 0.321}$   & \\\cline{2-7}
            & Total Time (s) &  $302.928 \pm 18.549$  &  $264.001 \pm 60.654$  &  $44.005 \pm 2.432$  &  $18.324 \pm 0.406$  &  $\bm{11.454 \pm 0.338}$   \\\cline{2-7}
        \hline\hline
        \multirow{6}{*}{25,000}
            & RMSE &  $1.096 \pm 0.141$  &  $1.075 \pm 0.176$  &  $5.523 \pm 13.421$  &  $0.907 \pm 0.269$  &  $\bm{0.892 \pm 0.190}$   \\\cline{2-7}
            & NAMI &  $0.129 \pm 0.035$  &  $0.146 \pm 0.042$  &  $0.145 \pm 0.059$  &  $0.109 \pm 0.136$  &  $\bm{0.277 \pm 0.084}$   \\\cline{2-7}
            & Forecast Horizon &  $60.5 \pm 35.4$  &  $79.8 \pm 49.4$  &  $73.8 \pm 73.5$  &  $86.0 \pm 76.1$  &  $\bm{228.1 \pm 56.9}$   \\\cline{2-7}
            & Train+Test Time (s) &  $23.367 \pm 4.677$  &  $20.160 \pm 13.237$  &  $2.047 \pm 0.128$  &  $\bm{1.155 \pm 0.304}$  &  $16.958 \pm 0.580$   \\\cline{2-7}
            & Tune Time (s) &  $312.270 \pm 18.715$  &  $304.800 \pm 59.900$  &  $58.783 \pm 3.346$  &  $\bm{24.209 \pm 1.033}$   & \\\cline{2-7}
            & Total Time (s) &  $335.637 \pm 20.387$  &  $324.960 \pm 69.449$  &  $60.830 \pm 3.391$  &  $25.364 \pm 1.163$  &  $\bm{16.958 \pm 0.580}$   \\\cline{2-7}
        \hline
    \end{tabular}
    \caption{Numerical experiment with varying training length and 20 total double-scroll independent trajectories. Each cell displays the mean metric across the 20 realizations, plus or minus their standard deviations. The bold cells represent the best model in the respective category. RMSE is Root Mean Square Error, NAMI is Normalized Average Mutual Information, and Forecast Horizon represents the number of time points for which the absolute residual error for all dimensions are less than the standard deviation of the training data for that respective coordinate. Forecast horizon represents how long the forecasts are qualitatively converged to the test data. Hyperparameter tuning for RNN, LSTM, RC, and NGRC uses TPE optimization (see the main manuscript) with 25 samples.}
    \label{tab:doublescroll_results}
\end{table}

\section{Van der Pol Oscillator}
Here we interrogate TreeDOX's ability to forecast a nonlinear system with stiff behavior and two disparate timescales. A prototypical example is the unforced Van der Pol oscillator \cite{cartwright1952van}:
\begin{align*}
    \dot{x} &= y \\
    \dot{y} &= -x + \mu (1-x^2)y
\end{align*}
where $\mu \geq 0$ is a damping parameter. The Van der Pol oscillator features a unique stable limit cycle for all $\mu\geq 0$, and becomes stiffer as $\mu$ grows. See Supplementary Fig.~\ref{fig:vdp_damp} to observe the effect of $\mu$ on Van der Pol dynamics. To test the effect of stiffness on forecasts, we vary $\mu$ on a uniform log scale from $\mu=1$ to $\mu=500$, and use RK45 to generate 50,000 and 1,500 training and testing points, respectively. Note that $dt$ is varied such that a uniform number of cycles is seen over the 1,500 test points. Supplementary Fig.~\ref{fig:vdp-forecasts} shows the forecast results, while Supplementary Fig.~\ref{fig:vdp-violin} summarizes the RMSE of the forecasts.

\begin{figure}
    \centering
    \includegraphics[width=0.95\linewidth]{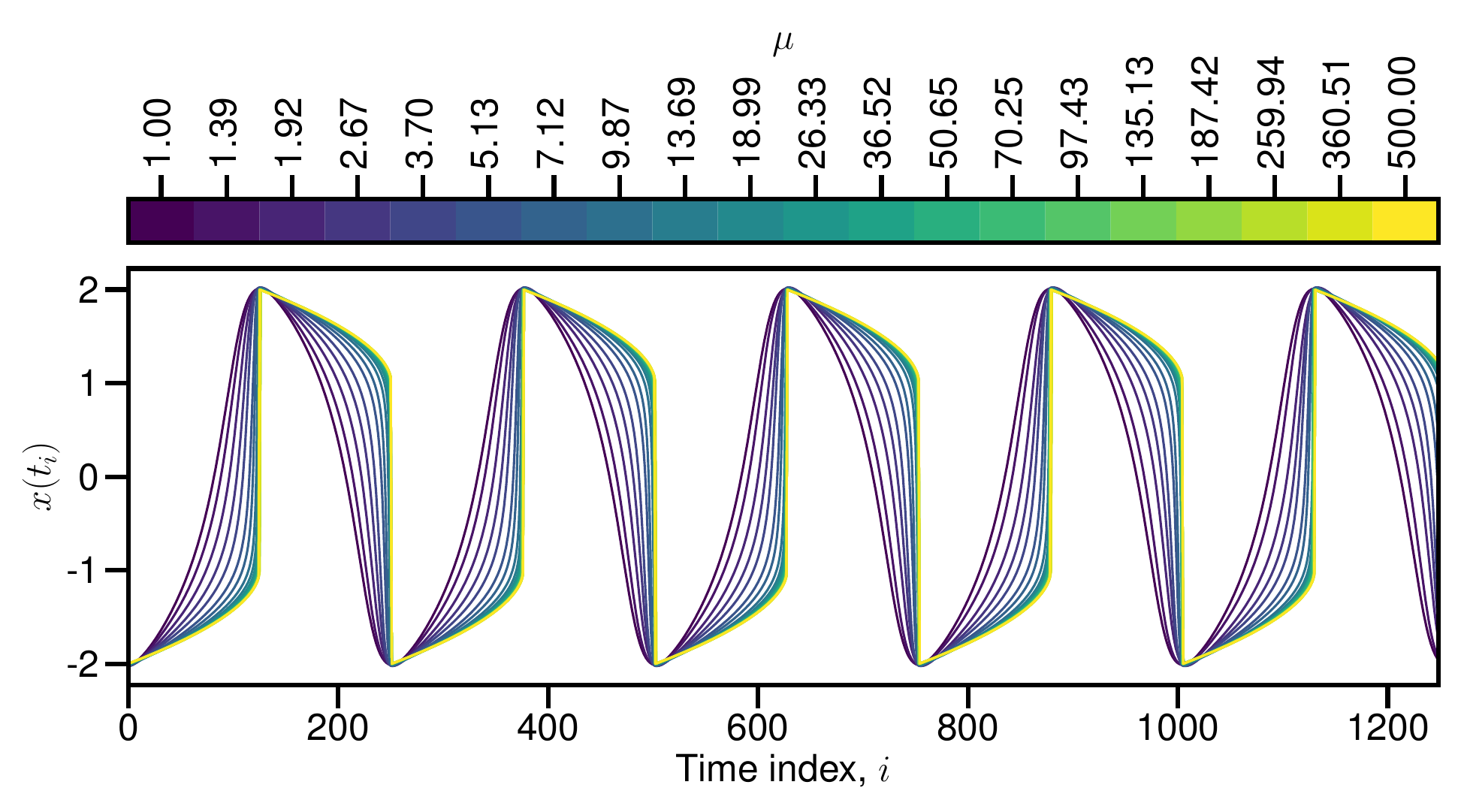}
    \caption{Van der Pol oscillator as the damping coefficient, $\mu$, varies. Here, the time axis of each curve is scaled to normalize the period between $\mu$ values. Observe the stiffening dynamics as $\mu$ grows.}
    \label{fig:vdp_damp}
\end{figure}

\begin{figure}
    \centering
    \includegraphics[width=0.95\linewidth]{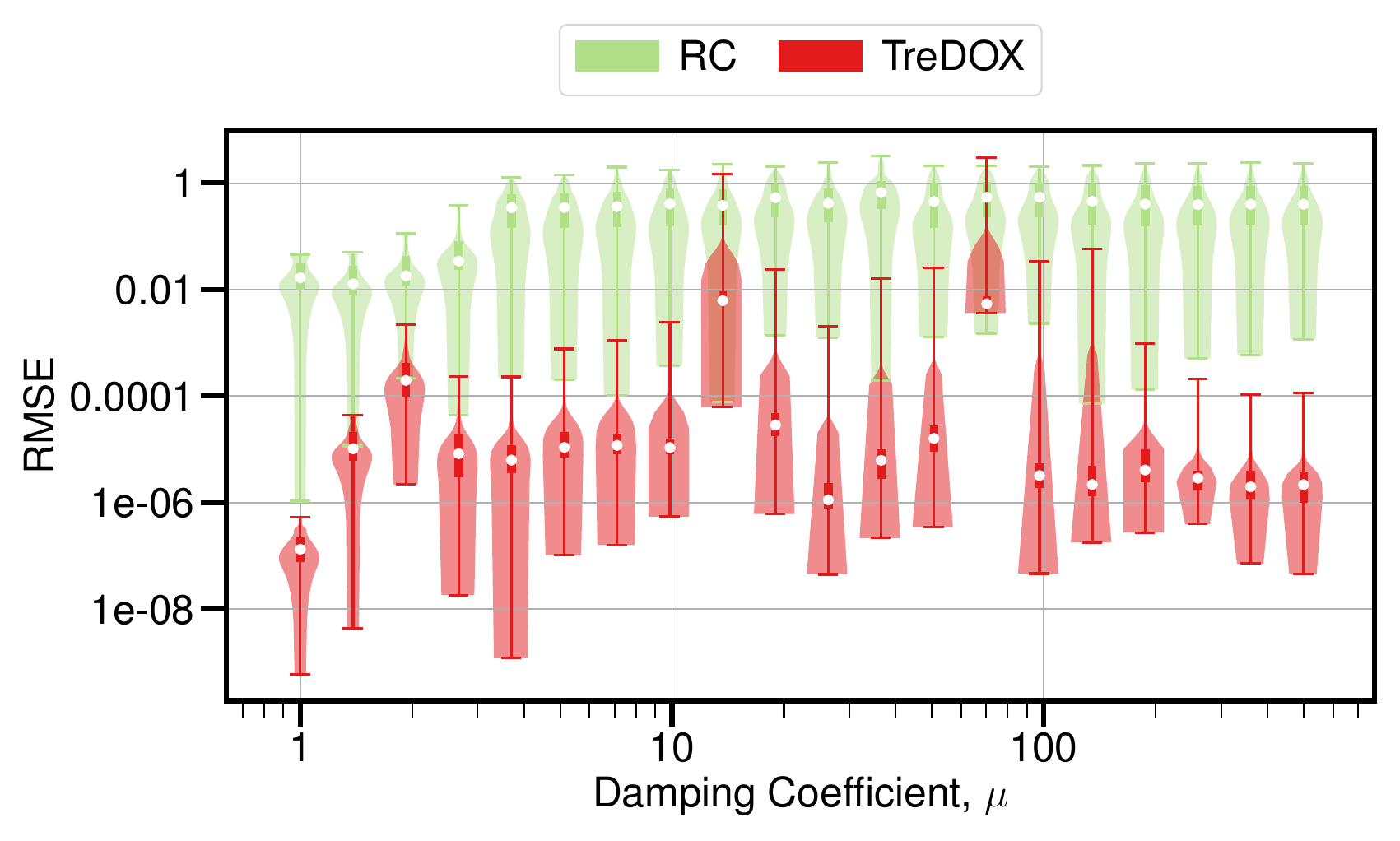}
    \caption{Violin plot of Root Mean Square Error (RMSE) values for RC and TreeDOX predictions (light green and red, respectively) of Van der Pol dynamics as the damping coefficient, $\mu$, varies. The lines, boxes, and white dot portray the bounds, 25th and 75th percentiles, and median, respectively.}
    \label{fig:vdp-violin}
\end{figure}

\begin{figure}
    \centering
    \includegraphics[width=0.9\linewidth]{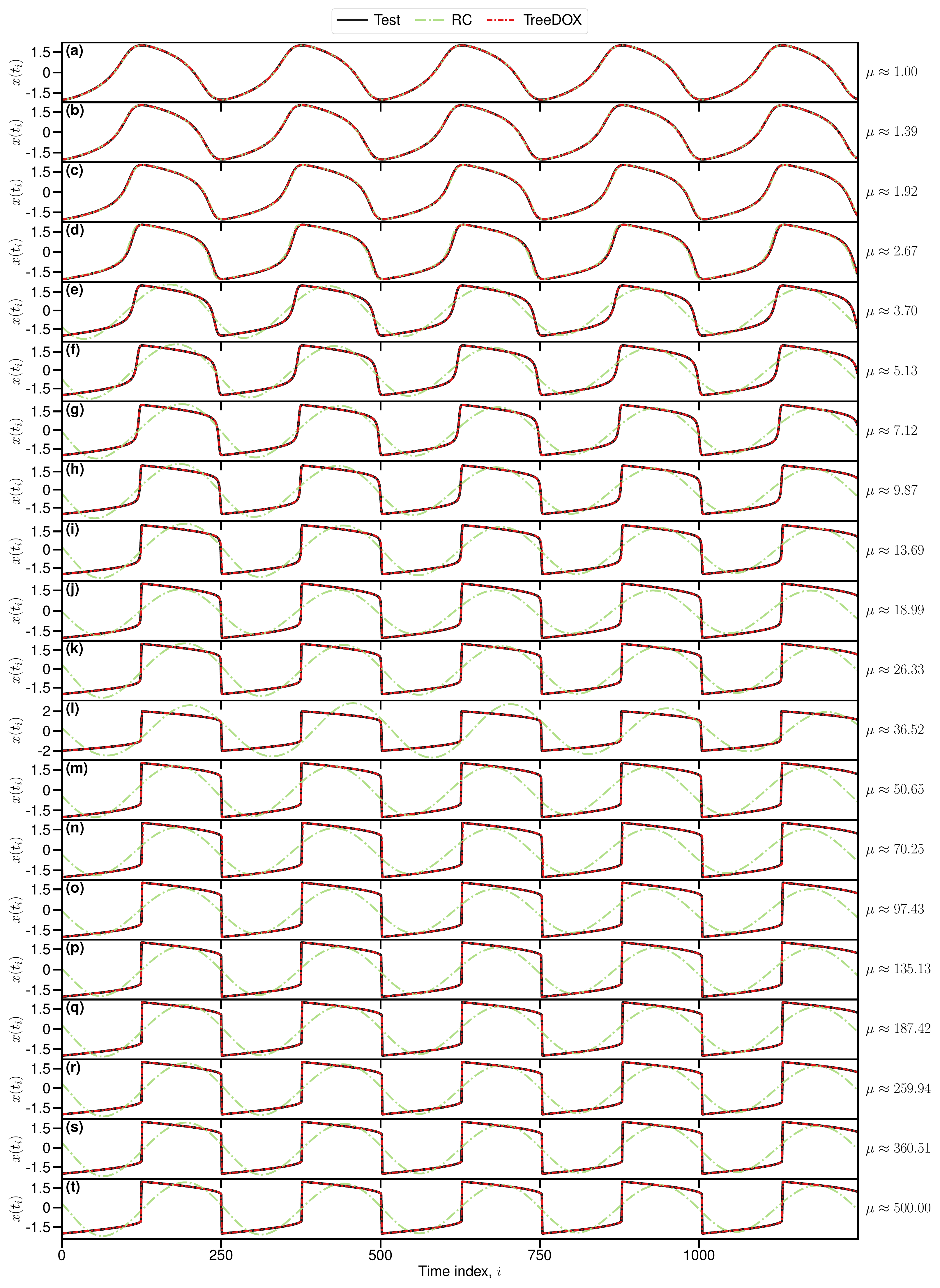}
    \caption{Van der Pol forecasts as the damping coefficient, $\mu$ varies. The time axis of each curve is scaled to normalize the period between $\mu$ values. Test data is black, while RC and TreeDOX are light green and red, respectively. Here, 50,000 and 1,500 training and testing points are used, respectively.}
    \label{fig:vdp-forecasts}
\end{figure}

\section{Lorenz96 System}
Here, we wish to investigate the effect of dimension on TreeDOX accuracy and speed. We use the Lorenz96 system, which features a variable dimension, $D$, allowing for direct scalability experiments \cite{lorenz1996predictability}. The governing equation is $\dot{x_i} = \left(x_{i+1}-x_{i-2}\right)x_{i-1} - x_i + F$ where $1\leq i \leq D$ is the $i$-th coordinate in the $D$-dimensional system, $F$ is a constant forcing value and $D\geq 4$. In this notation, $x_{-1}=x_{D-1}$, $x_0=x_D$, and $x_{D+1}=x_1$. See Supplementary Fig.~\ref{fig:lorenz96_d=5} and Supplementary Fig.~\ref{fig:lorenz96_d=10} for example forecasts for $D=5$ and $D=10$, respectively. $F=8$ is used for both examples, and the system is evolved with RK45 and $dt=0.01$ until $t_{max}=250$ and $t_{max}=500$, respectively. The last 1/16$^{th}$ time points for $D=5$ and 1/32$^{nd}$ time points for $D=10$ are held off as test data in both cases, producing 1,561 test points for both examples and 23,437 and 48,437 training points, respectively. For both cases, $x_1(t=0)=F+0.01$ and $x_i(t=0)=F$ for $i \in \{2,3,\dots,D\}$.

Lastly, we perform a more rigorous numerical experiment of the effect of dimensionality by varying $D$ from 4 to 10, and generating 20 realizations of Lorenz96 trajectories for each $D$ by varying $x_1(t=0)=F+0.01m$, where $m$ is the index of the respective realization. For each dimension and realization, we vary the amount of training data seen by the models. The test data remains fixed with a total length of 1,500 points, and the training length varies from 2,500 to 25,000. $F=8$ is used for $D\geq 5$, but $F=12$ is used for $D=4$ to ensure the dynamics remain chaotic. Note that here we exclude RNN and LSTM in our experiment in the interest of time, but keep RC and NGRC with automated hyperparameter tuning (TPE method as described in the main manuscript) with 25 samples. See Supplementary Fig.~\ref{fig:lorenz96} for a summary of results from the numerical experiment. Indeed, the speed of TreeDOX scales poorly for high dimensions when compared with RC or NGRC, but its accuracy remains competitive.

\begin{figure}
    \centering
    \includegraphics[width=0.8\linewidth]{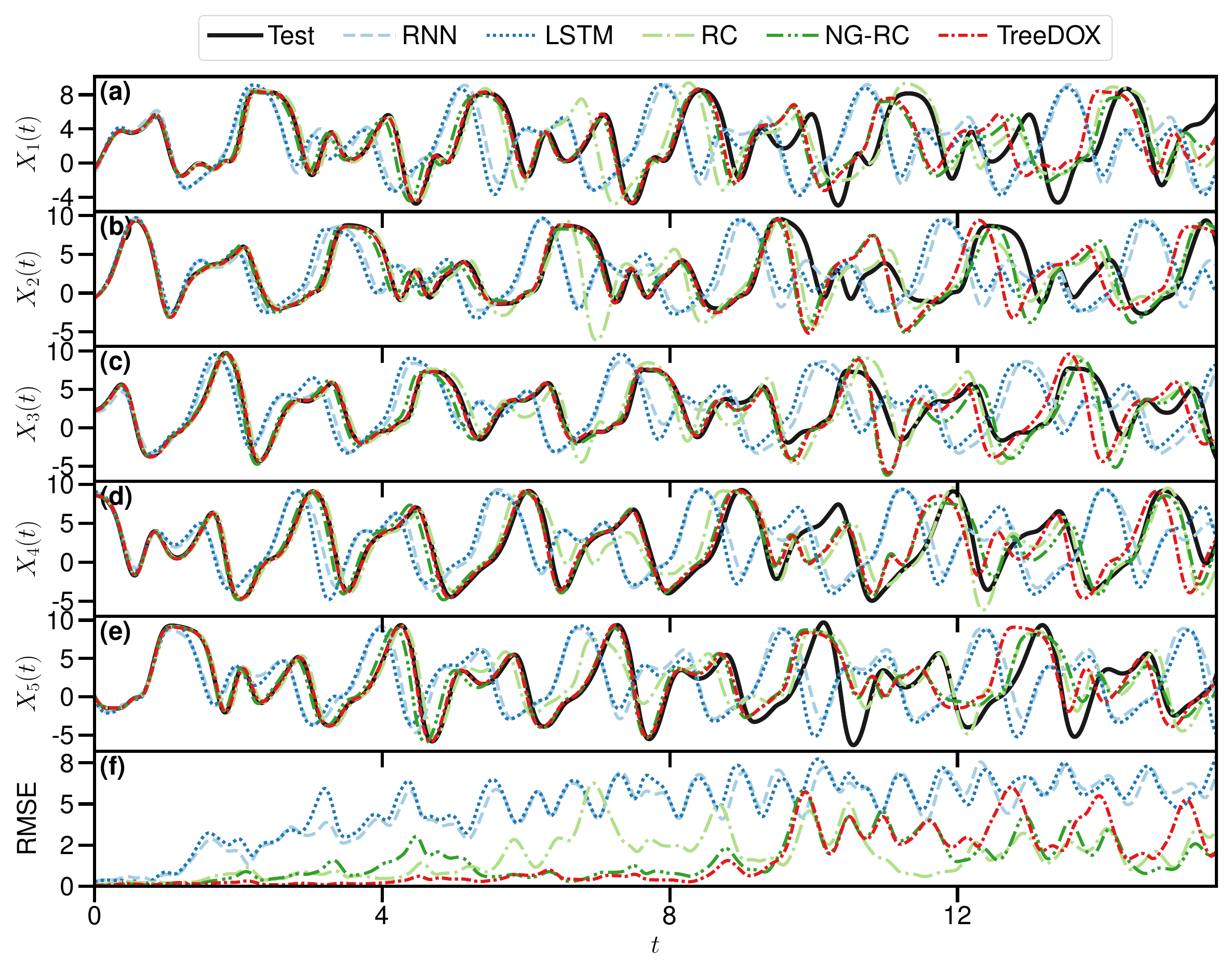}
    \caption{Example forecasts of Lorenz96 system with dimension $D=5$ and forcing $F=8$. Test data is shown in black, while RNN, LSTM, RC, NGRC, and TreeDOX forecasts are shown in light blue, blue, light green, green, and red, respectively. Here, 23,437 and 1,561 training and testing points are used, respectively.}
    \label{fig:lorenz96_d=5}
\end{figure}

\begin{figure}
    \centering
    \includegraphics[width=0.9\linewidth]{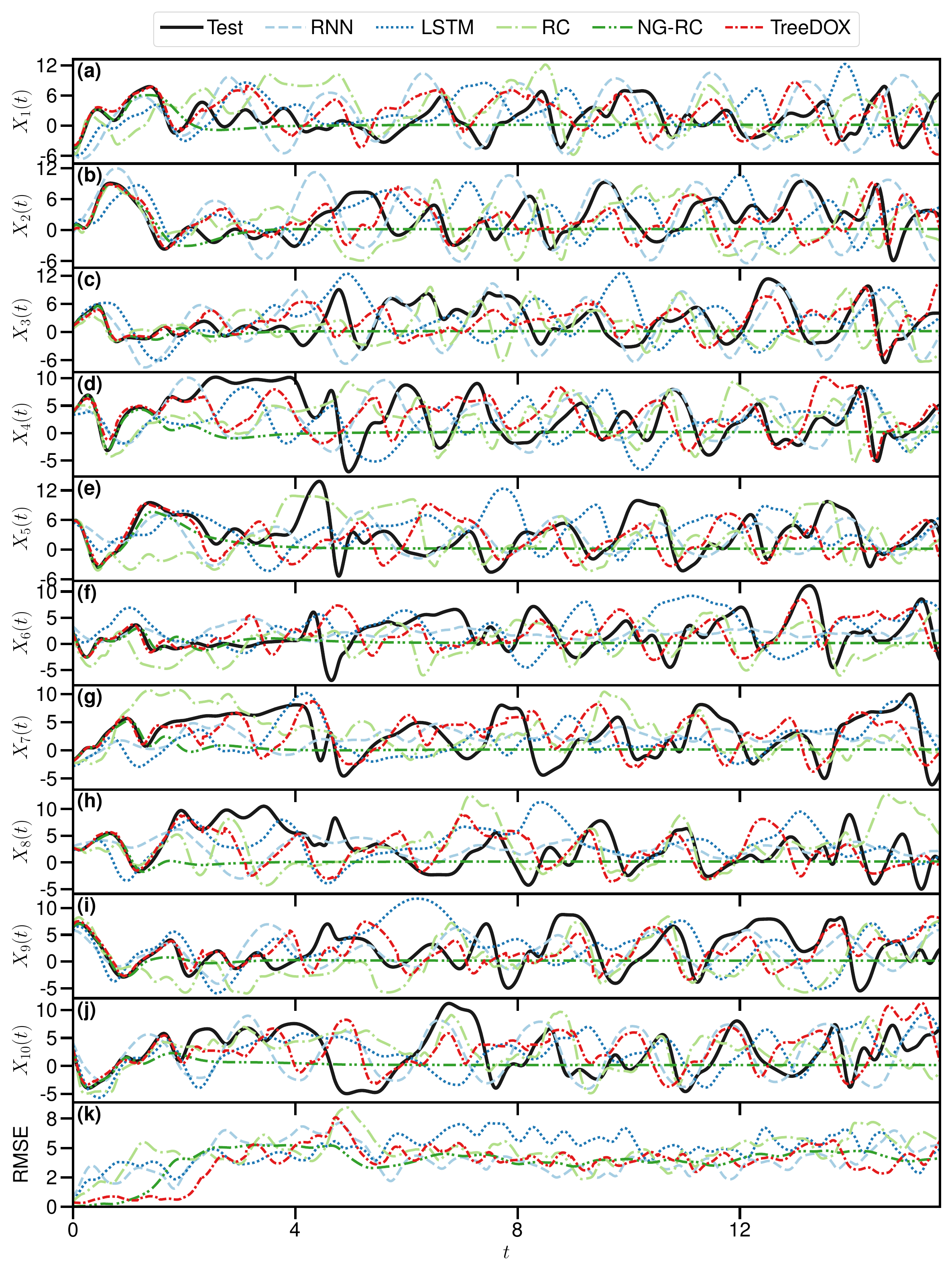}
    \caption{Example forecasts of Lorenz96 system with dimension $D=10$ and forcing $F=8$. Test data is shown in black, while RNN, LSTM, RC, NGRC, and TreeDOX forecasts are shown in light blue, blue, light green, green, and red, respectively. Here, 48,437 and 1,561 training and testing points are used, respectively.}
    \label{fig:lorenz96_d=10}
\end{figure}

\begin{figure}
    \centering
    \includegraphics[width=\linewidth]{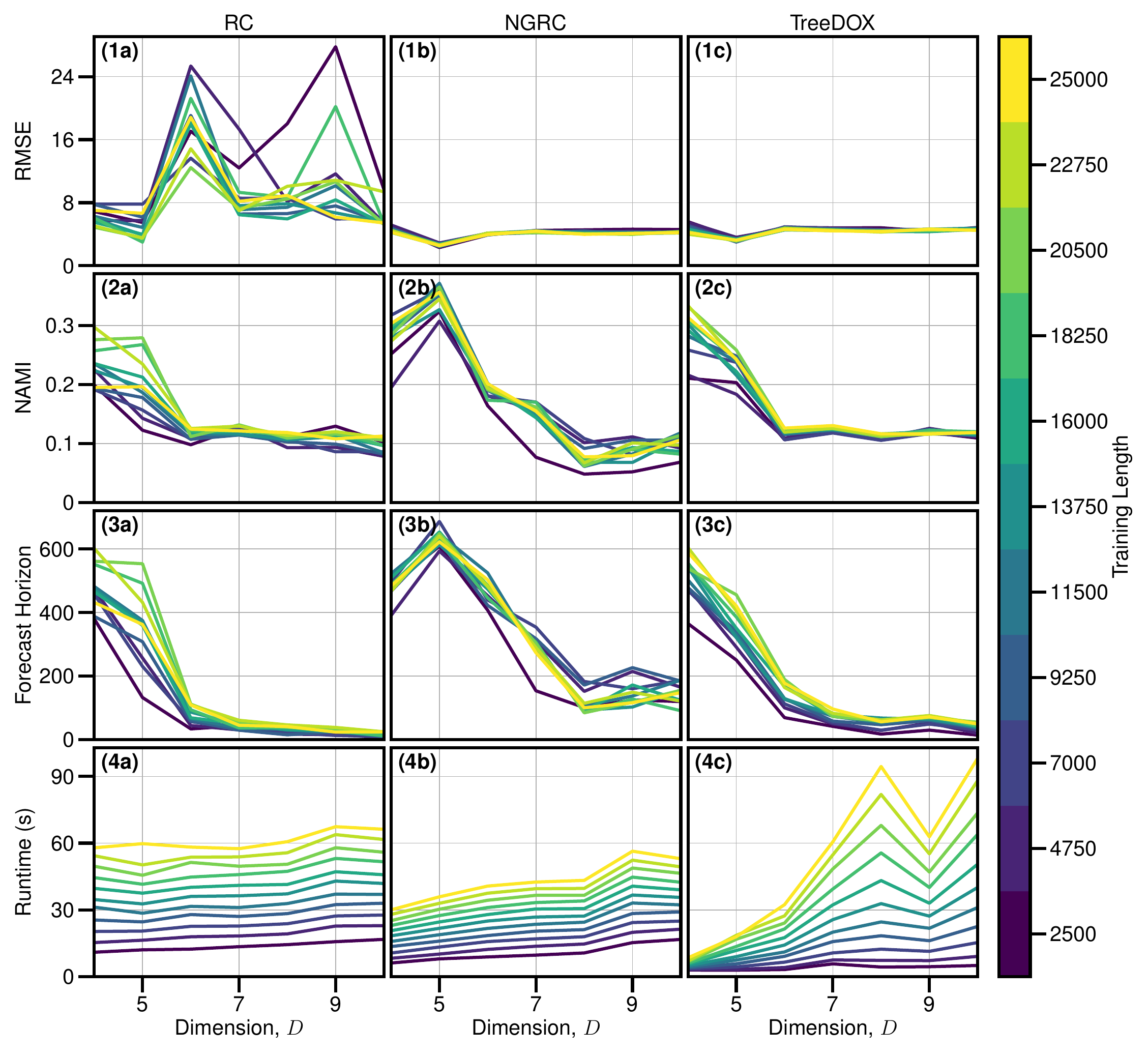}
    \caption{Dimension and training length experiment on the Lorenz96 system. For each dimension, 20 realizations of data are generated by perturbing initial conditions, then each model is fed a specified length of training data and asked to forecast the same test data. Each curve shown is the mean of the respective metric across all 20 realizations for the given dimension, where color indicates the length of training data.}
    \label{fig:lorenz96}
\end{figure}

\section{Discussion of GPU Parallelization}
In the \texttt{RAPIDS} ecosystem, \texttt{cuML}'s \texttt{RandomForestRegressor} implements RFs with GPU-parallelization, which provides substantial speedups compared to \texttt{Scikit-learn}'s \texttt{RandomForestRegressor} or \texttt{ExtraTreesRegressor} for users matching the software and hardware requirements \cite{RAPIDS, raschka2020machine, scikit-learn}. We perform a numerical experiment to investigate the performance gain when implementing \texttt{CuML}'s \texttt{RandomForestRegressor} in comparison to \texttt{Scikit-learn}'s \texttt{ExtraTreesRegressor} used in the rest of the results. Here, we generate 5 Lorenz realizations by perturbing the initial conditions: $(x(t=0),y(t=0),z(t=0)=(1+0.01m,1+0.01m,1+0.01m)$ where $m$ is the index of the realization. Then, for each realization, both TreeDOX implementations are fed a specified length of training data (up to 200,000 training points), and the speed and accuracy metrics are compared. At the time of writing, \texttt{CuML}'s \texttt{RandomForestRegressor} has two limitations compared to \texttt{Scikit-learn}'s \texttt{ExtraTreesRegressor}: first, only one output is permitted, meaning for each dimension in the data, a separate model must be trained; secondly, Gini feature importance is not implemented. In this experiment, feature importances for \texttt{CuML}'s \texttt{RandomForestRegressor} are calculated by first serializing the model to JSON format, then performing text searches to find the gain associated with each feature. See Supplementary Fig.~\ref{fig:CuML} for results of the experiment. Note in Supplementary Fig.~\ref{fig:CuML}(2a) the large additional time necessary to compute the feature importances. However, once feature importance is implemented in \texttt{CuML}'s \texttt{RandomForestRegressor}, this additional time would not be present, and using \texttt{CuML}'s \texttt{RandomForestRegressor} would, in fact, provide speedup over \texttt{Scikit-learn}'s \texttt{ExtraTreesRegressor}.

\begin{figure}[H]
    \centering
    \includegraphics[width=\linewidth]{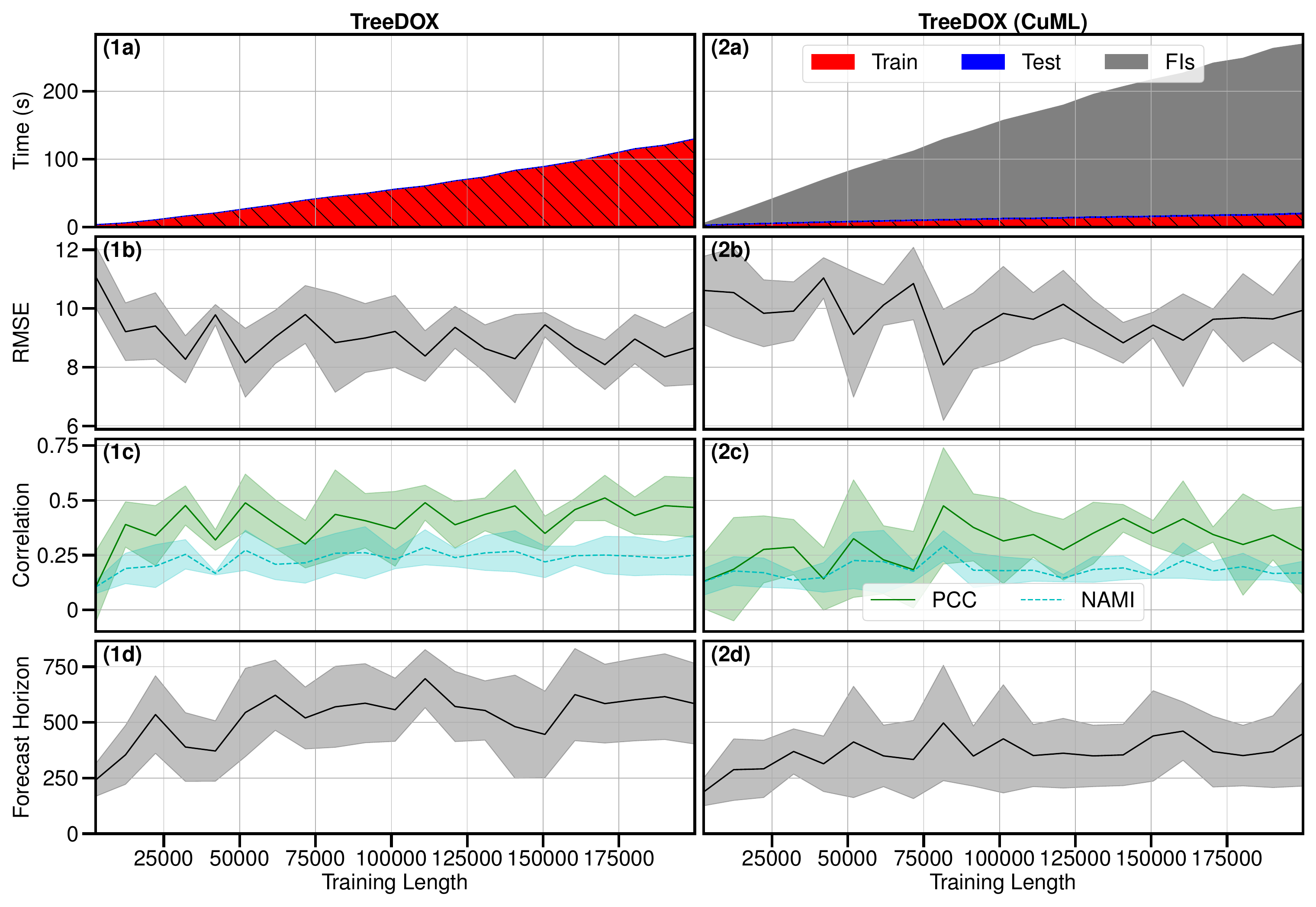}
    \caption{Numerical experiment investigating the speed gains while using \texttt{CuML}'s \texttt{RandomForestRegressor} rather than \texttt{Scikit-learn}'s \texttt{ExtraTreesRegressor} for Lorenz system data. Curves show the mean of 5 realizations and shaded areas show plus or minus one standard deviation.}
    \label{fig:CuML}
\end{figure}


\section{Implementations of Other Methods}

\subsection{Recurrent Neural Networks (RNN)}
Recurrent Neural Networks (RNN) are a variant of Artificial Neural Networks (ANN) which feature feedback loops within the hidden network, which provide the ability to model short-term memory of sequential data \cite{rumelhart1986learning}. We implement an RNN via Pytorch, using a delay embedding to improve training stability (the RNN is responsible for predicting the next point of all delay states, so the overlap of predictions and ground truth helps the model learn) \cite{paszke2019pytorch}. The hyperparameters (namely the learning rate, batch size, dropout, hidden network size, and warmup) were tuned manually to achieve the best possible performance on held-off validation data. Here, the hyperparameters tuned involve the delay embedding dimension, the number of training epochs, batch size, warmup length, dropout (only in the training phase), the number of layers, hidden network size, and the base learning rate (which is reduced when a plateau in validation loss is observed).

\subsection{Long Short-Term Memory (LSTM)}
Similar to RNN, Long Short-Term Memory networks (LSTM) feature a recurrent hidden network, but feature gating to help forget memory of long-past states \cite{hochreiter1997long}. We also implement LSTM in Pytorch, using the same delay embedding technique described above \cite{paszke2019pytorch}. Likewise, the hyperparameters were manually tuned. Here, the hyperparameters tuned involve the delay embedding dimension, the number of training epochs, batch size, warmup length, dropout (only in the training phase), the number of layers, hidden network size, and the base learning rate (which is reduced when a plateau in validation loss is observed).

\subsection{Reservoir Computing (RC)}
Reservoir Computing (RC) uses a similar framework to RNN, with the main difference being the complete randomization of the hidden network (called the reservoir) and training only being performed on the readout (output) layer \cite{jaeger2004harnessing}. We use the Python package ReservoirPy for our implementation of RC. While we attempted to implement hyperparameter optimization via the canned random search approach, we again opted to tune hyperparameters manually. Here, the hyperparameters tuned involve the number of units in the reservoir, the proportion of possible reservoir edges to include, the proportion of input-to-reservoir edges to include, the proportion of feedback (output-to-reservoir) edges to include, the scaling factor of feedback edges, the scaling factor of input-to-reservoir edges, warmup length, spectral radius of the reservoir adjacency matrix, leak rate, and ridge regression regularization factor.

\subsection{Next Generation Reservoir Computing (NG-RC)}
The recently developed Next Generation Reservoir Computing (NG-RC) converts RC into an equivalent Nonlinear Vector Autoregression machine (NVAR), which eliminates some hyperparameters and the need for a randomized reservoir \cite{gauthier2021next}. Here, we use ReservoirPy's NVAR implementation. Like the other models, we manually tune the hyperparameters (namely, the number of delay states and the ridge regression parameter). Here, the hyperparameters tuned involve the number of delay states, ridge regression regularization factor, warmup length, the highest polynomial order of nonlinear readout terms, and delay stride.

\begin{table}
    \centering
    \begin{tabular}{|c|c|c|c|}
        \hline
        RNN & LSTM & RC & NGRC \\
        \hline
        Embedding dimension & Embedding dimension & Reservoir units & Delay states \\
        Training epochs & Training epochs & Reservoir edges & Regularization \\
        Batch size & Batch size & Input edges & Warmup \\
        Warmup & Warmup & Feedback edges & Order \\
        Dropout & Dropout & Feedback scaling & Delay stride \\
        \# Layers & \# Layers & Input scaling &  \\
        Hidden size & Hidden size & Warmup &  \\
        Base learning rate & Base learning rate & Spectral radius &  \\
         &  & Leak rate &  \\
         &  & Regularization &  \\
        \hline
    \end{tabular}
    \caption{Model hyperparameters included in tuning.}
    \label{tab:hyperparams}
\end{table}

\end{document}